\documentclass[conference]{IEEEtran}
\IEEEoverridecommandlockouts
\usepackage{cite}
\usepackage{xurl}
\usepackage{float}
\usepackage{amsmath,amssymb,amsfonts}
\usepackage{algorithmic}
\usepackage{graphicx}
\usepackage{minted}
\usepackage{textcomp}
\usepackage{listings}
\makeatletter
\@ifundefined{iflatexml}{%
  \newif\iflatexml
  \latexmlfalse
}{}
\makeatother
\usepackage{xcolor}
\usepackage{fancyvrb}
\lstdefinestyle{jsonblock}{
  basicstyle=\fontsize{8.5pt}{10pt}\usefont{OT1}{cmtt}{m}{n},
  frame=single,
  framerule=0.4pt,
  framesep=8pt,
  rulecolor=\color{black},
  backgroundcolor=\color{white},
  breaklines=true,
  columns=fullflexible,
  keepspaces=true,
  showstringspaces=false,
  numbers=none,
  xleftmargin=0pt,
  xrightmargin=0pt,
  aboveskip=0.8em,
  belowskip=0.8em
}
\usepackage{tabularx} 
\usepackage{array}
\usepackage{booktabs}
\usepackage{placeins}
\usepackage{xcolor}
\usepackage{enumitem}
\usepackage[
  colorlinks=true,
  linkcolor=blue,
  citecolor=blue,
  urlcolor=blue
]{hyperref}
\usepackage{url}
\usepackage{titlesec}
\makeatletter
\def\@IEEEsectpunct{}
\@ifundefined{iflatexml}{%
  \newif\iflatexml
  \latexmlfalse
}{}

\makeatother

\def\BibTeX{{\rm B\kern-.05em{\sc i\kern-.025em b}\kern-.08em
    T\kern-.1667em\lower.7ex\hbox{E}\kern-.125emX}}

\newcommand{\CommunicationProtocolsTable}{%
\begin{table*}[t]
\centering
\caption{Communication protocols supported by TessIndex}
\label{tab:tessindex-communication-protocols}
\scriptsize
\renewcommand{\arraystretch}{1.18}
\begin{tabularx}{\textwidth}{
    >{\raggedright\arraybackslash}p{0.10\textwidth}
    >{\raggedright\arraybackslash}p{0.16\textwidth}
    >{\raggedright\arraybackslash}p{0.30\textwidth}
    >{\raggedright\arraybackslash}X
}
\toprule
\textbf{Protocol} & \textbf{Primary Object} & \textbf{Purpose} & \textbf{TessIndex Implementation} \\
\midrule

\texttt{A2A} &
Agents &
Standardized agent invocation, discovery, and agent to agent communication &
TessIndex exposes A2A compatible metadata, agent card references, and callable agent endpoints through the Primitive Wrapping Service. Native agent endpoints are mapped into A2A compatible request and response flows. \\

\midrule

\texttt{MCP} &
Tools and services &
Standardized tool and service invocation by agents &
TessIndex stores MCP server endpoints, tool metadata, health URLs, and capability descriptions so tools can be discovered and invoked through MCP compatible clients and orchestration kernels. \\

\bottomrule
\end{tabularx}
\end{table*}%
}
\newcommand{\CommerceProtocolsTable}{%
\begin{table*}[t]
\centering
\caption{Commerce protocols supported by TessIndex}
\label{tab:tessindex-commerce-protocols}
\scriptsize
\renewcommand{\arraystretch}{1.18}
\begin{tabularx}{\textwidth}{
    >{\raggedright\arraybackslash}p{0.10\textwidth}
    >{\raggedright\arraybackslash}p{0.22\textwidth}
    >{\raggedright\arraybackslash}p{0.28\textwidth}
    >{\raggedright\arraybackslash}X
}
\toprule
\textbf{Protocol} & \textbf{Role} & \textbf{Commerce Surface} & \textbf{TessIndex Implementation} \\
\midrule

\texttt{x402} &
Payment enabled agentic access &
Supports price discovery, payment aware requests, and paid task execution &
TessIndex stores price endpoints, task endpoints, wallet metadata, payment rail information, and settlement context so agents and services can participate in x402 compatible payment flows. \\

\midrule

\texttt{AP2} &
Authorization and mandate based commerce &
Supports explicit user authorization, mandate generation, and auditable commerce execution &
TessIndex exposes authorization metadata, payment context, audit trail references, and execution evidence needed to route agentic commerce through AP2 compatible workflows. \\

\bottomrule
\end{tabularx}
\end{table*}%
}

\iflatexml
\usepackage{latexml}
\fi

\begin{document}

\title{TessIndex: Capability Verified Identity System for the Agent Economy}

\author{
    \IEEEauthorblockN{
        % AUTHORS
        Mehul Goenka\textsuperscript{1,3}, 
        Tejas Pathak\textsuperscript{2,3}, 
        Siddharth Asthana\textsuperscript{1,3}
        \\[0.2em] % <--- This controls the vertical gap (adjust 1em as needed)
        % AFFILIATIONS
        \normalfont\small % Force text back to normal weight and smaller size   
        \textsuperscript{}                    \\
        \textsuperscript{1}University of Oxford\\
        \textsuperscript{2}Indian Institute of Technology, Delhi\\
        \textsuperscript{3}Tesseris.org
    }
}

\maketitle

\begin{abstract}
Software systems have traditionally been organized around applications where human users act as principal decision-makers. Recent developments in agentic capabilities alter this paradigm: software agents now autonomously translate high-level goals into structured tasks, orchestrating tools, services and sub-agents to execute complex workflows. This evolution gives rise to an agent economy where these autonomous agents capture real economic value. However, the infrastructure required to support the agent economy fails across three critical dimensions: the absence of persistent identity infrastructure prevents systemic accountability in agentic workflows; capability claims remain self-declared not backed by verifiable execution evidence; and the disconnect between creator identities, agent performance, and project value hinders the economic valuation of agents as assets. While existing registries provide naming and discovery, unifying these features around a persistent identity anchor remains largely unaddressed.

TessIndex is a capability-verified identity system for agent primitives that utilizes a dual-plane architecture: the blockchain records compact commitments for identity, ownership, and verification, while centralized servers maintain dynamic metadata for discovery, commerce, and reputation. It establishes: persistent identities across agent primitives to enforce systemic accountability in autonomous workflows; a predicate-based verification process replacing self-declared claims with cryptographic capability proof; an identity infrastructure that links agent performance to both project and creator identities while capturing value through tokenization. Ultimately, TessIndex serves as an integrated infrastructure that binds an agent’s existence across capabilities, execution, and reputation into a single persistent identity.  

\end{abstract}


\section{Introduction}
Software systems have traditionally been organized around applications in which human users select interfaces, authorize actions, and remain the principal decision-makers. Recent developments in agent capabilities shift this interaction model by driving the transition toward an intent-driven economy. Instead of relying on human intervention, modern agents can interpret complex user intents, dynamically orchestrate across external agents and specialized tools, and autonomously drive end-to-end execution and settlement. These capabilities create the foundation of the agent economy: a digital environment where these autonomous agents and their supporting tools coordinate to execute workflows that produce economic value.

The agent economy is not composed of agents alone. An agent may depend on tools for deterministic functions, skills for reusable instructions, services for external commercial capabilities, and communication interfaces for invocation and settlement. We use the term agent economy primitive to refer to these independently identifiable components.

As workflows become more autonomous, each primitive must be discoverable, and its endpoints must be resolvable. However, discovery alone is insufficient. Before a user, agent, or an orchestration system delegates work or value, it must be able to answer three foundational questions: Who is acting and under whose authority? What are they demonstrably capable of doing? Can the resulting interaction, payment, and reputation be reconstructed across systems?

Currently, the agent economy lacks a shared accountability infrastructure capable of bridging human intent and verified capabilities with discovery, payment, reputation, and dispute resolution across organizational boundaries. This deficit manifests across three core gaps:
\begin{itemize}
    \item \textit{Lack of systemic accountability:} Agentic workflows rely on top-down delegation, where humans set an overarching intent and agents invoke system primitives to fulfil that intent. However, accountability breaks down because current identity infrastructure cannot trace the chain of authority from humans, through agents, down to the invoked primitives. As a result, current systems fail to bind underlying actions to the governing human intent that authorized them.

    \item \textit{Unverified claims of agent capability:} Current systems depend on unverified, self-asserted capability claims for both agents and primitives. As such, these claims lack both systematic verification and concrete execution evidence under their declared operating conditions. Ideally, concrete evidence must capture what a primitive can demonstrably do under the set conditions, along with its actual execution record.

    \item \textit{Missing creator, performance, and value link:} Agents coordinate sub-agents, tools, and primitives into structured projects that drive autonomous execution, building reputation and economic value through sustained market interaction. Current infrastructure, however, cannot reliably link agent performance to creator identities and a project's economic value. Bridging this gap requires a foundational identity layer that defines concrete performance metrics and establishes mechanisms for value capture and distribution.

\end{itemize}

Prior work on agent identity and registries spans both research and industry. Research efforts~\cite{agentRegistrySurvey} such as the NANDA Index~\cite{nandaIndex} and Agent Name Service~\cite{agentNameService} focus on naming, discovery, metadata resolution, and secure interaction. However, their scope remains largely limited to agents. They do not extend identity to other core components such as tools and services, nor do they address the economic layer of the agent economy. Industry efforts extend this direction through marketplaces, reputation systems, validation records, and token based agent markets. Examples include ERC 8004~\cite{erc8004}, Fetch.ai~\cite{fetchai}, Virtuals Protocol~\cite{virtualsProtocol}, HOL Registry~\cite{holRegistry}, and HCS 14~\cite{hcs14}. However, these systems still address specific parts of the identity problem. None provides a unified identity architecture that spans the components, trust requirements, and economic dimensions of the broader agent economy.

This paper introduces TessIndex, a capability verified identity system for the agent economy. TessIndex is built on a dual plane architecture that persists identity across the blockchain and a centralized registry. Through this architecture, TessIndex addresses three foundational gaps across identity, trust, and economy. Under identity, it provides a persistent identity for agents, tools, skills, and services. Under trust, it verifies their claimed capabilities and security posture, with the resulting proofs recorded on the blockchain. Under economy, it immutably links commercial utility and economic reputation to each persistent identity, while storing the associated metadata in the registry.

\subsection*{\textbf{Contributions}}
TessIndex makes the following key contributions:
\begin{itemize}
    \item First, it introduces a unified dual plane identity system for agent economy primitives, with immutable artifacts stored on the blockchain and richer metadata maintained in the registry.
   
    \item Second, it introduces a capability verification system that evaluates the claimed capabilities and security posture of each primitive, with the resulting cryptographic proof stored on the blockchain.

    \item Third, it defines a rich registry schema for commercial and performance-related metadata, enabling primitives to be easily discovered, evaluated, and connected to their persistent identity.

    \item Fourth, it introduces an interoperability framework that supports multiple protocols and standards, together with a federated trust model for cross registry identity mappings.
    
\end{itemize}

The remainder of this paper is organized as follows. We first review prior work on agent identity, discovery, and emerging industry standards to identify the limitations of existing approaches and derive the requirements for a broader identity system for the agent economy. We then define the design goals of TessIndex and present its overall architecture, including the dual plane structure and six core system domains. Each domain is subsequently examined in detail through the fields and subfields governing identity, control, interoperability, commerce, assetization, and trust. Finally, we describe the TessIndex process flow across registration, verification, listing, and lifecycle management, before concluding with directions for future work.

\section{Background and Related Work}

Prior work around agent identity and registries can be broadly divided into two categories. The first category consists of research and academic work that studies agent discovery, naming, registry design, verifiable metadata, and secure resolution. The second category consists of industry systems and standards that are being deployed as agent marketplaces, agent registries, communication protocols, and commerce layers.

\subsection{Research and Academic Related Work}

A closely related line of research is the NANDA Index, which proposes infrastructure for discoverability, identity, and authentication in the Internet of AI Agents. NANDA introduces a lean index that resolves to dynamic AgentFacts, where AgentFacts act as cryptographically verifiable metadata records containing information needed for secure agent discovery and interaction. While NANDA addresses how agents can be found, authenticated, and resolved across a large network, it does not fully define a broader identity record for agent economy primitives that includes ownership binding, record authority, wallet binding, commerce metadata, assetization metadata, payment context, reputation aggregation, task verification history, and human readable trust representation.

Agent Name Service, or ANS, is another closely related academic proposal. ANS introduces a DNS inspired directory for secure AI agent discovery and interoperability. It uses public key infrastructure, registration and renewal, capability aware resolution, structured communication metadata, and protocol adapters for agent communication systems such as A2A~\cite{a2aSpec}, MCP~\cite{mcpSpec}, and ACP~\cite{acpSpec}. However, ANS remains primarily a naming and secure discovery layer. It gives agents resolvable names and supports secure interaction, but its core function remains closer to directory infrastructure than comprehensive identity record.

\subsection{Industry Related Work}

On the industrial side, ERC 8004 proposes trustless agents through three registries: an Identity Registry, a Reputation Registry, and a Validation Registry. This directly aligns with the need for decentralized agent identity, feedback, and validation. However, ERC 8004 remains lightweight and registry specific, and therefore does not incorporate the broader components required for supporting commerce and trust.

Fetch.ai and Agentverse provide industry infrastructure for deploying, discovering, and connecting autonomous agents~\cite{fetchai,agentverse}. The Fetch ecosystem uses the Almanac~\cite{fetchaiAlmanac} as a registry component that allows access to registered agents and related information, while Agentverse provides discovery and marketplace style visibility for hosted agents. However, this infrastructure does not define a general purpose, ecosystem neutral identity record for agent economy primitives across many registries, chains, marketplaces, and protocols.

Virtuals Protocol approaches the agent economy from the direction of tokenization, ownership, and market formation~\cite{virtualsProtocol}. It assumes the existence of reliable identity, capability, performance, and trust records, but does not itself solve these underlying problems.

HOL Registry and HCS 14 focus on universal agent infrastructure, cross protocol indexing, routing, and portable identifiers~\cite{holRegistry}. HOL describes its Registry Broker as a universal index and routing layer for AI agents, while HCS 14 defines a Universal Agent ID using decentralized identifier concepts across web APIs, web3 protocols, and hybrid systems. While these systems address universal indexing, routing, and cross registry portability, they do not fully define the complete identity record required for agentic commerce.

\subsection{Gap Analysis}

The above analysis highlights three core gaps:

\begin{itemize} 
\item \textbf{First,} existing identity systems do not cover the full set of primitives that constitute the agent economy. Most systems are designed around agents alone, while the agent economy also depends on tools, services, skills, endpoints, wallets, domains, protocols, verification modules, and other functional primitives. As a result, the identity layer remains incomplete.
\item \textbf{Second,} 
existing systems do not cover the full identity lifecycle and interaction surface required for agentic systems. Some systems focus on naming, others on discovery, others on communication, routing, reputation, tokenization, or registry anchoring. However, no existing identity system unifies naming, discovery, communication, routing, ownership, control, endpoint resolution, commerce metadata, reputation, and verification into a single coherent identity architecture. 
\item \textbf{Third,} existing systems do not provide a sufficiently robust and verifiable method for validating capability claims. Many systems allow agents to publish capability descriptions or service metadata, but these claims are often self declared, weakly verified, or disconnected from execution evidence. For an agent economy to function reliably, capability claims must be supported by verifiable attestations, task evidence, and reputation signals. 
\end{itemize}

\section{Design Goals}

\subsection{Unified Identity System for Primitives in Agent Economy}
This design goal builds on the fragmentation described in the previous section, where primitives within the agent economy currently exist across disconnected identity systems. In this context, primitives refer to the core constituents of the agent economy, including agents, tools, services, and other components that collectively sustain agentic activity. For the agent economy to thrive, there is a need for a common and unified identity system that can represent these primitives consistently. Such a system can reduce fragmentation, enable composability, and unlock stronger network effects across the broader agent economy.

\subsection{Stable and Persistent Identity for Trust in Commerce}
This design goal addresses one of the foundational requirements for trust in agentic commerce: stable and persistent identity. Any user interacting with an agent, tool, or service should be able to verify both the authenticity of the primitive and the authenticity of its author or creator. A persistent identity enables users to build confidence over time, while allowing agents and tools to accumulate reputation, transaction history, and commercial credibility. This, in turn, can increase adoption, transaction volume, and trust in agent mediated commercial operations.

\subsection{Verification of Claimed Capabilities}
This design goal addresses another core dimension of trust: the verification of capability claims. An agent may claim to perform a specific function, or a tool may claim to satisfy certain security properties, but such claims create adoption friction when they are not supported by verifiable proof. To address this, primitives should be subject to appropriate verification processes, with the resulting verification receipts anchored in an immutable and tamper resistant manner. This allows users, agents, and orchestration systems to rely on verified claims rather than unsupported assertions.

\subsection{Intent-Based Discovery with Easy Integration}
This design goal focuses on reducing friction in discovery and adoption. The system should support intent based discovery by LLMs, orchestrator agents such as Claude Code and Codex CLI, and full fledged agentic kernels. It should expose rich metadata that allows the right primitive, whether an agent, tool, or service, to be selected efficiently based on context, capability, interface, and verification status. The discovery experience should also support easy integration across multiple user journeys, including interactive interfaces, generative UI, CLI based workflows, MCP integration, and simple API endpoint integration.

\subsection{Interoperability with External Ecosystems and Protocols}
This design goal emphasizes the need for an identity system that can interoperate with external ecosystems and protocols. As discussed in the previous section, several identity systems and protocol frameworks already exist, including NANDA Index, HOL Registry, Google A2A, and Coinbase x402~\cite{x402Spec}. A robust identity layer for the agent economy should therefore not operate in isolation. It should be capable of interfacing with existing registries, protocols, and discovery systems so that primitives can participate across ecosystems while retaining a stable, verifiable  and transferable identity.

% ----------------------------------------------------
% SECTION IV
% ----------------------------------------------------
\section{Architecture}

% Placeholder for Figure 1 if you wish to insert it here
\begin{figure*}[htbp]
    \centering
    \includegraphics[width=\textwidth]{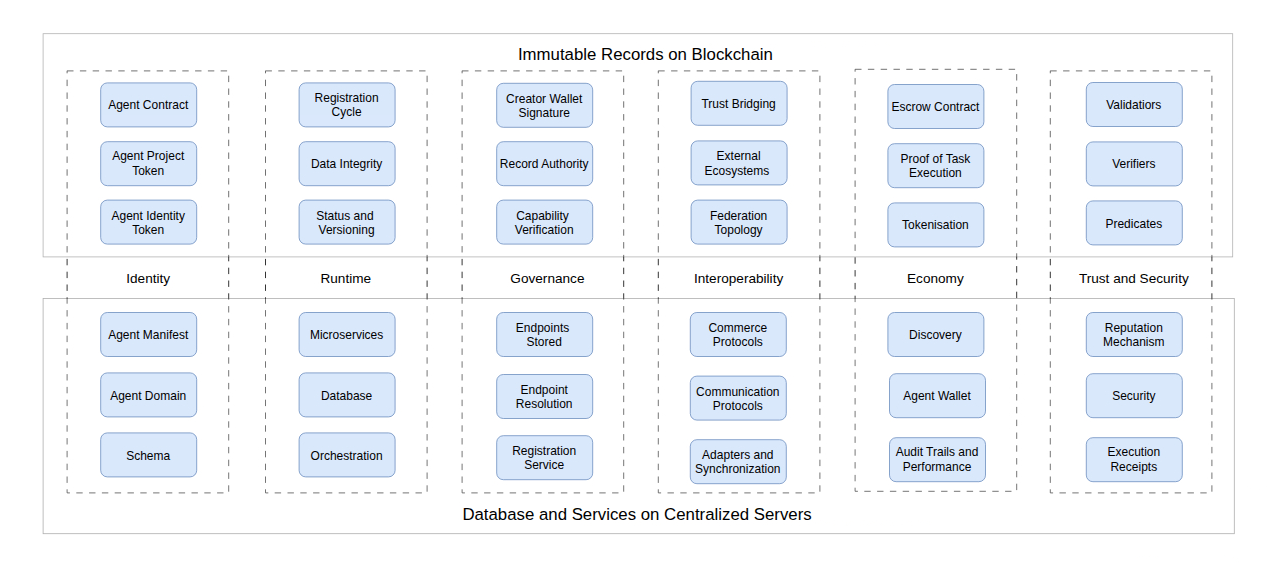} % Note: Changed linewidth to textwidth
    \caption{\textbf{TessIndex two plane architecture across six key domains:} Overview of the TessIndex architecture, illustrating the separation between the blockchain (anchoring identity, contracts, verification, escrow, tokenization, and validator functions) and centralized servers (managing manifests, domains, schemas, orchestration, endpoints, discovery, reputation, and execution records). The entire system is mapped across six core domains: Identity, Runtime, Governance, Interoperability, Economy, and Trust and Security.}
    \label{fig1}
\end{figure*}

To satisfy the stated design goals, we propose the following hybrid architecture which can be visualized across two orthogonal dimensions:

\subsection{Two Plane Model}

To effectively balance decentralization with performance, the TessIndex architecture operates across a dual-plane model, distributing data and services depending on their functional purpose (see Fig. 1):
\begin{enumerate}[label=\arabic*)]
    \item Immutable Records on Blockchain
    \item Database and Services on Centralized Servers
\end{enumerate}

\vspace{1em}
The blockchain provides trust, security, and verifiability, while a centralized database provides flexibility, scalability, and metadata expressiveness. This separation allows TessIndex to achieve a practical trade off between the complexity of identity metadata and the security guarantees required for a trustworthy identity system.

\subsubsection{Immutable Records on Blockchain}

The blockchain provides trust and security by anchoring cryptographic commitments to the metadata stored in the centralized database. This plane stores compact and verifiable representations such as hashes, state commitments, ownership records, and update histories (\textit{see Fig. 1, Immutable Records on Blockchain}). These records act as immutable proofs of the existence, integrity, and evolution of identity related metadata.

By placing only security critical information on chain, TessIndex can benefit from the transparency, auditability, and consensus guarantees of blockchain infrastructure without incurring the cost and scalability limitations of storing large or frequently changing metadata directly on chain. The blockchain therefore serves as the root of trust for the system. Any metadata stored in the centralized database can be independently verified by comparing it against the corresponding commitment on blockchain.

\subsubsection{Database and Services on Centralized Servers}

The centralized servers store and manage the richer metadata required by the identity system. Specifically, this includes identity descriptors, metadata, service endpoints, commercial data, schema definitions, and other contextual attributes that may be too large, dynamic, or privacy sensitive to store directly on the blockchain (\textit{see Fig. 1, Database and Services on Centralized Servers})

This plane is designed for flexibility and efficient retrieval. It enables TessIndex to support complex identity representations, updates, discovery, and application specific extensions while keeping the footprint on blockchain minimal. Since the integrity of primitive metadata is anchored through the blockchain, the system can maintain verifiability even when metadata is stored outside the blockchain. Sensitive information can also remain encrypted or selectively disclosed, allowing the architecture to balance transparency with privacy.

\subsection{Six Key Domains}

Orthogonal to the two storage planes, TessIndex is organized across six key domains (\textit{see Fig. 1}). Each domain addresses a specific set of concerns required to support a complete identity infrastructure. These domains are not isolated components, but cooperating abstractions that together define how identities are created, controlled, used, exchanged, and trusted. The domains are described as follows.

\subsubsection{Identity}

The Identity domain (\textit{see Fig. 1, Identity}) serves as the foundational naming and classification system for the ecosystem. It defines the complete scope of supported entities and assigns them stable, canonical identifiers bound to explicit ownership. Additionally, it establishes the standardized schemas and comprehensive metadata required to consistently represent these entities, their capabilities, and their commercial attributes across different platforms.

\subsubsection{Runtime}

The Runtime domain (\textit{see Fig. 1, Runtime}) defines the internal operating model through which TessIndex registers, verifies, resolves, updates, and revokes primitives to maintain a deterministic global state. It covers the core services, data model, orchestration layer, kernel level execution flow, memory structure, policy checks, and lifecycle mechanisms that coordinate registry activity across internal and on chain environments to ensure idempotent state machine transitions.

\subsubsection{Governance}

The Governance domain (\textit{see Fig. 1, Governance}) provides the operational and technical infrastructure necessary to manage the entire lifecycle of indexed entities. It houses the core microservices for naming, capability verification, and registration, while enforcing strict record authority and authenticated updates. Furthermore, this layer handles the storage and dynamic, constraint based resolution of endpoints, ensuring a seamless developer experience through automated publishing pipelines.

\subsubsection{Interoperability}

The Interoperability domain (\textit{see Fig. 1, Interoperability}) bridges the gap between fragmented ecosystems by enabling unified communication, discovery, and commerce across external ecosystems. It integrates industry standard protocols for both agentic interaction, such as A2A and MCP, and payments, while managing a federated networking topology. Through dedicated adapters and cross registry identity mapping, it ensures that entities can be universally discovered and invoked while preserving their original provenance and trust signals.

\subsubsection{Economy}

The Economy domain (\textit{see Fig. 1, Economy}) powers the economic activity and user centric discovery within the agent economy. It facilitates intent based search and sophisticated retrieval mechanisms, securely binds autonomous wallets to agent identities, and orchestrates end to end, verify then pay settlement flows. Beyond transactional payments, it also oversees the tokenization lifecycle, providing the infrastructure for creators to launch, govern, and trade agent linked assets.

\subsubsection{Trust and Security}

The Trust and Security domain (\textit{see Fig. 1, Trust and Security}) establishes the cryptographic and operational guarantees required for reliable agentic interactions to prevent unauthorized state transitions. It enforces a comprehensive verification process for both agent capabilities and task execution, relying on on chain trust anchors and trusted execution environments to validate all cryptographic task commitments. Supported by strict access controls, abuse resistance mechanisms, and a dual plane reputation system, it ensures that all activity in workflows remains secure, factual, and strictly accountable.

\vspace{\baselineskip}

% ----------------------------------------------------
% SUB-SUBSECTION 1
% ----------------------------------------------------
\section{Key Domains}

\subsection{Identity}

The Identity domain \textit{(see Fig. 2)} defines how TessIndex determines what is indexed, how indexed entities are identified, and who is authorized to mutate records over time. It establishes the internal object model and the interoperability surface across external ecosystems, anchors agent identity through a stable canonical identifier with on-chain root of trust, and specifies the authority and audit semantics for record updates. Together, these controls ensure that discovery remains consistent across heterogeneous ecosystems, identity remains verifiable and portable for commerce workflows, and record mutation remains deterministic, authenticated, and traceable.

\begin{figure*}[htbp] 
    \centering 
    \includegraphics[width=\textwidth]{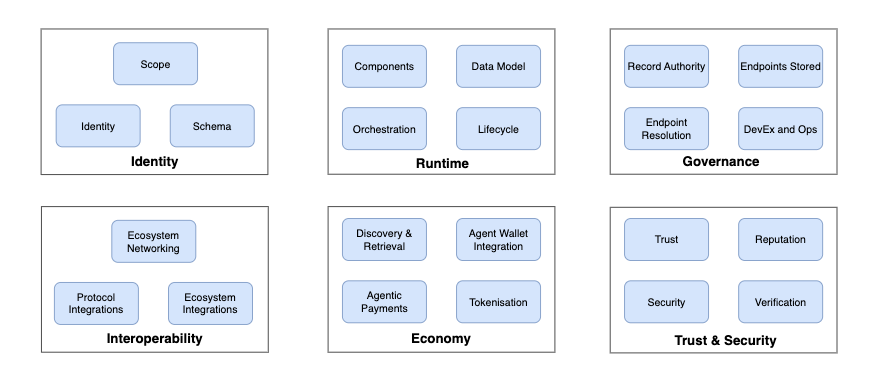} 
    \caption{\textbf{Six key domains with associated fields:} Organization of the registry by Identity, Runtime, Governance, Interoperability, Economy, and Trust and Security across naming, control services, protocol support, discovery, payment, tokenization, verification, lifecycle management, and auditability.} 
    \label{fig:tessindex_domains}
\end{figure*}

\subsubsection{Scope}

This field defines the types of entities that TessIndex stores and the ecosystems it can support for discovery and commerce. It also clarifies how ecosystem integration is managed at the identity layer. This field therefore defines the intended coverage boundary of TessIndex and frames ecosystem integration through the lens of identity.

\paragraph{Objects}
TessIndex stores several entity types as first class records in the registry. This ensures clarity and consistent semantics across ecosystems during discovery and integration. It also supports standardization by characterizing each entity, defining its schema, creating adapters, and specifying integration points across ecosystems. TessIndex supports registration of the following primary object types:

\begin{itemize}
    \item \textbf{Agents:} Any AI entity, whether a standalone agent or a multi agent system, equipped with a resolvable endpoint for communication. In this work, an agent is assumed to execute a reasoning loop over user intent using a large language model while invoking integrated tools and services to generate an output. TessIndex remains model agnostic and does not constrain the internal implementation of the agent.

    \item \textbf{Tools:} Any capability that can be invoked by an agent as part of its workflow. Tools are typically hosted as MCP servers and invoked through API calls. In this work, tools are treated as deterministic components that process information and produce outputs in a predictable manner, unlike agents. Examples include a Google Search tool or a GitHub MCP server.

    \item \textbf{Services:} Any non agentic SaaS system, catalog, or technical infrastructure that uses agents to drive economic activity. Services are typically integrated with agents for discovery and payment settlement. Examples include an e commerce catalog seeking agentic commerce support or an investment model integrated with a portfolio management agent.

    \item \textbf{Skills:} Modular capabilities that enhance an agent's functionality. Each skill packages instructions, metadata, and optional resources such as scripts or templates that an agent uses depending on user intent. Skills are added to an agent's working repository as folders and contain a \texttt{SKILL.md} file that captures metadata such as the skill name and description. Examples include a Canvas Design skill or an Excel skill.
     
    \item \textbf{Channels:} Messaging platforms and other communication surfaces through which a user interacts with an AI agent. Examples include WhatsApp, Telegram, Discord, and Slack.

    \item \textbf{Projects:} Any group of agents organized together based on a shared scope, purpose, or brand. A creator or organization can organize agents and primitives into projects to reflect a collective purpose or brand for assetization. For example, a productivity suite may contain a meeting scheduler, an automated mailing agent, and a document writing agent.
    
    \item \textbf{Agentic Apps:} Applications that are interactive and generative in terms of user interface and user experience. They support bidirectional state changes and can adjust their interface based on user intent and conversation context. They are usually integrated with an agent  or an orchestration model that acts as the non deterministic reasoning component.
\end{itemize}

TessIndex models explicit relationships between these objects and supports basic lifecycle states for each record, including \texttt{unverified}, \texttt{inactive}, \texttt{active}, \texttt{verified}, and \texttt{revoked}. These lifecycle states preserve discovery semantics over time.

\paragraph{Ecosystem Scope}
This field defines the ecosystems and interoperability surfaces that TessIndex supports from a discovery and commercial perspective. It also explains how connectivity and information management are handled across these ecosystems. This clarifies the coverage and authority of records, including whether a record is native to TessIndex or derived from an external registry, and how provenance, freshness expectations, and conflicts should be interpreted.

TessIndex is designed to be interoperable across ecosystems for both discovery and agentic commerce. For discovery, it maintains a \texttt{source\_id} and an adapter for each external registry. Each adapter defines a clear ingestion interface and a mapping mode, such as mirroring a record or maintaining a reference to it. The adapter also contains the logic required to preserve the provenance of external records in a verifiable manner and to deterministically handle collisions across sources through an explicit precedence policy.
\begin{figure*}[t]
    \centering
    \includegraphics[width=0.85\linewidth]{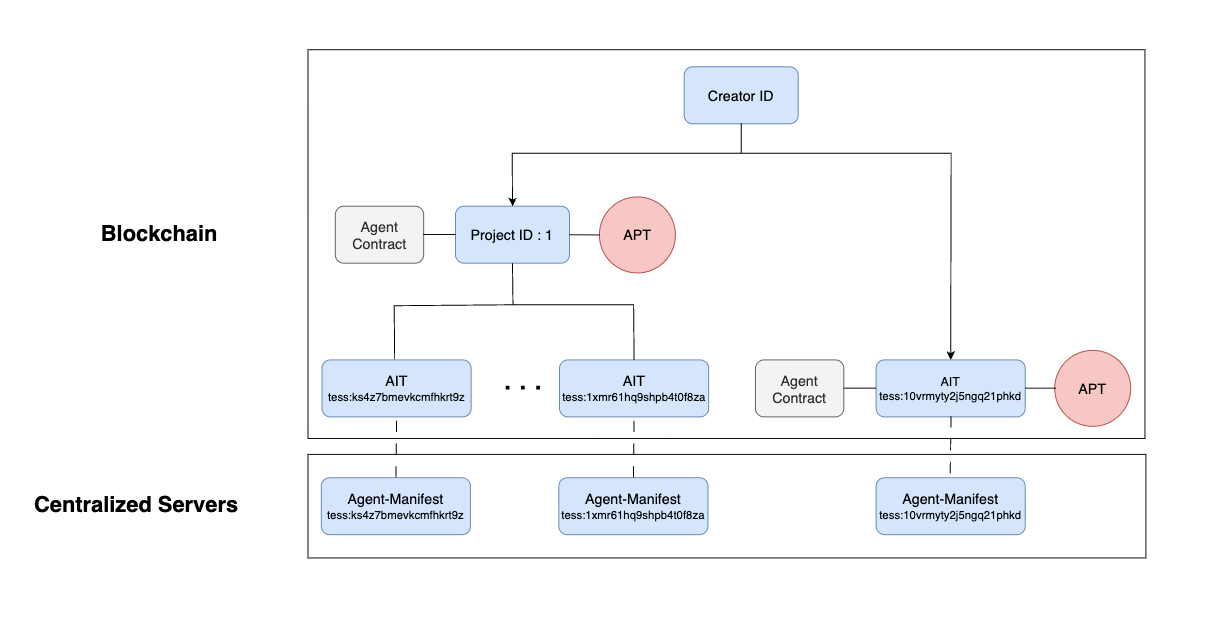}
    \caption{\textbf{Identity Structure:} Relationship between Agent Manifest, Projects, Agent Identity Tokens, Agent Contracts, and Agent Project Tokens}
    \label{fig:agent-identity-relationship}
\end{figure*}
\subsubsection{Identity}

Identity defines the identifier surfaces supported by TessIndex, the canonical identity anchor used for record linkage, and the binding between an agent and an accountable owner. It also specifies the human facing identity surface and naming rules used for discovery.

\paragraph{Identifiers}
This field defines the core identity primitives for agents stored in the TessIndex registry. It describes the different types of identifiers and how they map across different facets of agentic commerce, allowing an agent to be discovered and referenced across registries and commerce flows. This is necessary because agentic commerce requires stable identity across multiple representations: immutable references for trust, ownership, and settlement, and identifiers for interoperability and human facing discovery.

Every agent is assigned a unique 96 bit cryptographically secure random identifier encoded in Crockford Base32, referred to as an \texttt{agent\_id}. An example is shown below:

\begin{verbatim}
agent_id = tess:0123456789abcdefghjk
\end{verbatim}

The \texttt{agent\_id} acts as the central identity anchor~\cite{didCore} for an agent across the ecosystem and can be expressed through multiple identifier surfaces depending on the integration context. Every agent has a dual plane identity: one in the centralized database and the other on blockchain. In the centralized database, agent metadata is stored in persistent MongoDB storage. The blockchain identity resides in an Agent Identity Token, or AIT, following the ERC 721~\cite{eip721} token standard. It is minted by the Agent Factory smart contract on blockchain. The AIT preserves the integrity of the metadata by storing a cryptographic hash of the same on chain. 

Agent identity can exist as a standalone record or be organized hierarchically under a \texttt{project\_id}. Each project contains an explicit on chain reference to the creator ID \textit{(see Fig. 3)}. Each project, or standalone agent, is assigned a unique Agent Contract that defines the verification predicates for every agent. Further, Agent Project Tokens, or APTs, are minted for every \texttt{project\_id}, while standalone agents may be represented directly through an AIT. These tokens support assetization of agentic intellectual property and may be traded against stablecoin cryptocurrencies, backed by the performance and utility metrics of the corresponding agents.

\paragraph{Ownership Binding}
This subfield defines what constitutes control over an identity and how TessIndex binds identity to an owner. It captures the mechanism used to establish and verify ownership and determines who is authorized to act on behalf of the agent in commerce and lifecycle operations. This is required to prevent impersonation and to ensure that payments, tokenization, and record mutation are consistently attributable to a verified controlling party.

TessIndex primarily binds an \texttt{agent\_id} to ownership through an Agent Identity Token and a \texttt{project\_id} when the agent is part of a project. The Agent Identity Token provides an immutable and verifiable anchor that links the agent to the creator ID and establishes the controlling authority for the agent's identity. Every Agent Identity Token minted under a \texttt{project\_id} explicitly references the owner in an immutable manner. This on chain ownership binding is designed to prevent impersonation and make identity portable across commerce primitives such as assetization and settlement without relying on off chain attestations.

\paragraph{Agent Domain Model}
This subfield defines both the human facing identity surface for an agent and the information consolidated for end users. It captures the role of an agent domain in supporting user centric discovery and describes the corresponding implementation structure.

Analogous to Web2 discovery mechanisms, each agent in the ecosystem is assigned an agent domain that acts as the user centric surface for the agent's identity and presence. TessIndex uses this domain to present a consolidated view of the agent as determined by the creator, including the human readable name, agent metadata, traction, performance metrics, and marketing information. The agent domain is defined by the Primitive Naming Service (\textit{see Components field in Runtime Domain}) and follows the format below:

\begin{lstlisting}[
  frame=single,
  keepspaces=true,
  showstringspaces=false,
  breaklines=true,
  tabsize=2
]
<primary_protocol>://<agent_name>.<agent_id>.<agent_capability>.v<version>
\end{lstlisting}

\begin{table*}[t]
\centering
\caption{Core schema domains in TessIndex}
\label{tab:tessindex-core-schema}
\scriptsize
\renewcommand{\arraystretch}{1.18}
\begin{tabularx}{\textwidth}{
    >{\raggedright\arraybackslash}p{0.12\textwidth}
    >{\raggedright\arraybackslash}p{0.22\textwidth}
    >{\raggedright\arraybackslash}p{0.25\textwidth}
    >{\raggedright\arraybackslash}X
}
\toprule
\textbf{Domain} & \textbf{What it Contains} & \textbf{Purpose} & \textbf{Core Fields} \\
\midrule

Identity &
Core agent identity, capability verification, and metadata for enabling universal discovery &
Create a universally resolvable and stable reference for every indexed object &
\texttt{agent\_id}, \texttt{agent\_name}, \texttt{ait\_ref}, \texttt{root\_domain}, \texttt{schema\_version}, \texttt{skills[]}, \texttt{capabilities[]}, \texttt{input\_mode}, \texttt{output\_mode}, \texttt{status}, \texttt{created\_at}, \texttt{updated\_at} \\

\midrule

Control &
Ownership binding and update authorization &
Ensure only authorized signatures can create, update, or revoke records &
\texttt{creator\_id}, \texttt{auth\_scheme}, \texttt{eip712\_domain} \\

\midrule

Interoperability &
Support for external ecosystems with protocol identifiers &
Allow other agent networks to discover how to communicate with the indexed object &
\texttt{protocols[protocol\_id, external\_id, endpoints, core\_fields[]]} \\

\midrule

Commerce &
Data related to integrated services and commerce primitives such as the agent wallet address &
Support price discovery and payment routing without heavy metadata in the registry &
\texttt{services[service\_id, name, description, price, currency]}, \texttt{agent\_wallet\_addr}, \texttt{audit\_trail}, \texttt{predicate\_ids[]} \\

\midrule

Asset Linkage &
Off chain link to APT and related metadata, telemetry, and utility data &
Bind asset to blockchain identity and capture performance for fundamental analysis &
\texttt{apt\_ref}, \texttt{total\_workflows}, \texttt{avg\_success\_rate}, \texttt{tokens\_used}, \texttt{avg\_latency}, \texttt{avg\_throughput} \\

\bottomrule
\end{tabularx}
\end{table*}

The components of the agent domain are defined as follows:

\begin{itemize}
    \item \texttt{primary\_protocol}: Identifies the interaction protocol for the entity. It is \texttt{a2a} for agents and \texttt{mcp} for tools.

    \item \texttt{agent\_name}: Human readable agent name used for discoverability and user experience.

    \item \texttt{agent\_id}: The 96 bit Crockford Base32 identifier for the agent, serving as the central identity anchor across both on chain and off chain planes.

    \item \texttt{agent\_capability}: Encodes the capability and functions possessed by the agent, including integrated tools and services, for capability driven discovery.

    \item \texttt{version}: Agent version used for unambiguous resolution across evolving interfaces and schemas.
\end{itemize}

For example, a Portfolio Manager Agent with token swapping capability may be represented as follows:

\begin{lstlisting}[
  frame=single,
  keepspaces=true,
  showstringspaces=false,
  breaklines=true,
  tabsize=2
]
AgentName: Portfolio Manager Agent
agent_id: tess:dxc5f73847t001q228d
AgentDomain: a2a://portfolio-manager-agent.dxc5f73847t001q228d.token_swap.v0.0.1
\end{lstlisting}

\paragraph{Human Readable Names}
This subfield defines how TessIndex supports user friendly naming without compromising uniqueness and safety. It captures naming rules, uniqueness constraints, namespaces, and collision handling. This is needed because humans search by names, but names are also a primary spoofing surface. Without clear rules, discovery becomes ambiguous and attackers can exploit naming collisions.

TessIndex supports \texttt{AgentName} as the primary human readable identifier used by creators and end users for discovery. Names are subject to hash based uniqueness checks during registration to avoid collisions. They are also gated through capability verification, ensuring that names resolve to agents that meet defined verification thresholds. This reduces confusion and spoofing risk at the discovery layer.

\subsubsection{Schema}

The Schema field defines the structural contract TessIndex uses to represent agents, tools, and services in a manner that is interoperable by default and extensible by design. It establishes the minimum required record shape for baseline compatibility, the extension surface for richer discovery and operations, and the structured representation of capabilities so that verification and execution planning can be performed deterministically. It also specifies the commerce and assetization metadata required to support settlement and tokenization workflows natively at the centralized services.

\paragraph{Optional Fields}
This subfield defines the extension surface beyond the core schema. It captures additional non mandatory fields that enhance discovery, operations, and ecosystem specific interoperability without breaking clients that only implement the core schema. This is needed because different workflows require different levels of metadata depth, and TessIndex must support richer functionality while preserving a stable baseline.

TessIndex supports an optional field layer that can be included where available and safely ignored by clients that do not require it. These fields include details that are not part of the required core schema, such as marketing data, creator defined custom fields, extended discovery information, and ecosystem specific metadata.

\paragraph{Core Fields}
This subfield defines the minimum required fields for a valid TessIndex record. It captures the strict baseline schema that every compliant record must satisfy so that any resolver, orchestrator, or external ecosystem integration can interpret the record consistently. This is needed to guarantee interoperability even when clients only implement the minimum interface.

TessIndex therefore enforces a strict required core schema that is intentionally small and stable, forming the canonical valid record contract. The core fields span five key domains:

\begin{itemize} 
\item \textbf{Identity:} Core fields that are required to define identity, and support naming and discovery. 
\item \textbf{Control:} Fields required to define ownership, control, and update authorization of a primitive. 
\item \textbf{Interoperability:} Fields required to support baseline interoperability with external ecosystems. 
\item \textbf{Commerce:} Fields required to support commerce by integrating services and storing commerce primitives. 
\item \textbf{Asset Linkage:} Fields required to link tradable assets off chain with performance data for fundamental analysis. 
\end{itemize}

The Table~\ref{tab:tessindex-core-schema} outlines each of these domains and describes how TessIndex captures the corresponding core fields.

\paragraph{Schema Richness}
This subfield defines the breadth and depth of metadata TessIndex can express. It captures richer attributes used for higher order operations such as ranking, routing, compatibility checks, and cross registry normalization. This is needed to enable sophisticated discovery and execution behavior beyond simple lookup, and to align with the metadata surfaces exposed by other registries and catalogs.

TessIndex supports a detailed schema that incorporates the minimum amount of information required to support agentic commerce without introducing unnecessary complexity. This approach harmonizes fields commonly used across other indexing services, enabling richer interoperability and operational semantics. Rich schema fields at the level of agent identity, including description, skills, capabilities, example use cases, and status, support discovery by improving semantic retrieval and ranking. The interoperability layer adds further depth by including fields from different ecosystems, thereby supporting discovery and establishing trust in agentic commerce. The fields for commerce and asset linkage provide the core primitives for price discovery, commercial trust, and fundamental analysis of tradable agent linked assets.

\paragraph{Capability Definition}
This subfield defines how capability claims are represented so that they are both verifiable and execution relevant. It captures a structured capability representation, including explicit input and output formats, constraints, and representative subtasks, so that capability verification and orchestrator driven execution can be carried out deterministically rather than relying on free text descriptions. This is needed to reduce ambiguity in what an agent or tool can do and to support reliable matching between task intent and callable capabilities.

TessIndex only accepts capabilities with clearly defined interfaces, constraints, and representative subtasks. A well defined interface, such as JSON, text, or image input and output, enables easier integration of an agent into the system and supports capability verification. Clear definitions of constraints and representative subtasks also improve discovery by orchestrator agents, where agents with the most semantically matching subtasks may be given preference. Further, the capability definition includes the corresponding \texttt{predicate\_id} chosen from the predicate library for the specified capability. This \texttt{predicate\_id} with the corresponding verification mechanism is used for verifying the primitive's capability in a Trusted Execution Environment.

\paragraph{Agentic Commerce Metadata}
This subfield defines the commerce relevant attributes required to discover and execute paid services. It captures fields such as pricing signals, supported payment rails, settlement modes, wallet addresses, and service or capability listings necessary to route commerce workflows into authorization and settlement. This is needed to make commerce operations seamless by ensuring that discovery returns not only what can be done, but also how it can be paid for and settled.

TessIndex includes commerce related metadata for every agent to support price discovery and payment rail integration. Every agent is assigned a native wallet for receiving agentic payments in a chain agnostic manner. The schema stores the agent wallet address along with payment rail details to support settlement. The schema also stores metadata related to services integrated with the agent. This includes the registered service ID, request URLs, authentication method, payment mode, price, description, and capability information for the service. This provides a robust discovery surface for service providers seeking to enhance their reach through agentic commerce. Finally, the schema adds a layer of trust to agentic commerce by storing the audit trail URL along with verification predicate IDs for seamless verification. The following block defines an example schema for commerce-related metadata.

\iflatexml
\par
\begin{center}
\begin{minipage}{0.90\paperwidth}

\begin{lstlisting}[
  frame=single,
  showstringspaces=false,
  breaklines=true,
  escapeinside={(*@}{@*)}
]
(*@\makebox[\linewidth][l]{\texttt{\{}}@*)
(*@\hspace*{1.2em}@*)commerce: {
(*@\hspace*{2.4em}@*)agent_wallet: {
(*@\hspace*{3.6em}@*)address: 0x8f3C2A7d9E4B1c6F2aA91b44D7E3c4A9f1E2D3C4,
(*@\hspace*{3.6em}@*)payment_rail: {
(*@\hspace*{4.8em}@*)rail_id: rail_usdc_base,
(*@\hspace*{4.8em}@*)network: base,
(*@\hspace*{4.8em}@*)asset: USDC
(*@\hspace*{3.6em}@*)}
(*@\hspace*{2.4em}@*)},
(*@\hspace*{2.4em}@*)services: [
(*@\hspace*{3.6em}@*){
(*@\hspace*{4.8em}@*)service_id: svc7K2M9Q4T8X,
(*@\hspace*{4.8em}@*)service_name: RWA Compliance Check,
(*@\hspace*{4.8em}@*)description: Executes sanctions screening, issuer validation, and eligibility checks,
(*@\hspace*{4.8em}@*)request_urls: {
(*@\hspace*{6em}@*)request_url: https://agent.example/api/v1/compliance/check,
(*@\hspace*{6em}@*)quote_url: https://agent.example/api/v1/compliance/quote
(*@\hspace*{4.8em}@*)},
(*@\hspace*{4.8em}@*)authentication: {
(*@\hspace*{6em}@*)auth_method: OAuth2.0
(*@\hspace*{4.8em}@*)},
(*@\hspace*{4.8em}@*)payment: {
(*@\hspace*{6em}@*)payment_mode: pay_per_request,
(*@\hspace*{6em}@*)payment_details: {
(*@\hspace*{7.2em}@*)rail_id: rail_usdc_base,
(*@\hspace*{7.2em}@*)terms: prepaid
(*@\hspace*{6em}@*)}
(*@\hspace*{4.8em}@*)},
(*@\hspace*{4.8em}@*)price: {
(*@\hspace*{6em}@*)amount: 4.5,
(*@\hspace*{6em}@*)currency: USD,
(*@\hspace*{6em}@*)unit: per_check
(*@\hspace*{4.8em}@*)}
(*@\hspace*{3.6em}@*)}
(*@\hspace*{2.4em}@*)],
(*@\hspace*{2.4em}@*)audit_trail_url: https://agent.example/audit/agt_01jz7m4x6n8p2q9r3s5t7u9v,
(*@\hspace*{2.4em}@*)predicate_ids: [
(*@\hspace*{3.6em}@*)pred_payment_rail_verified_v1,
(*@\hspace*{3.6em}@*)pred_service_endpoint_healthcheck_v1,
(*@\hspace*{3.6em}@*)pred_compliance_attestation_v1
(*@\hspace*{2.4em}@*)]
(*@\hspace*{1.2em}@*)}
}
\end{lstlisting}

\end{minipage}
\end{center}
\par
\else

\begin{lstlisting}[
  frame=single,
  keepspaces=true,
  showstringspaces=false,
  breaklines=true,
  tabsize=2
]
{
  commerce: {
    agent_wallet: {
      address: 0x8f3C2A7d9E4B1c6F2aA91b44D7E3c4A9f1E2D3C4,
      payment_rail: {
        rail_id: rail_usdc_base,
        network: base,
        asset: USDC
      }
    },
    services: [
      {
        service_id: svc7K2M9Q4T8X,
        service_name: RWA Compliance Check,
        description: Executes sanctions screening, issuer validation, and eligibility checks,
        request_urls: {
          request_url: https://agent.example/api/v1/compliance/check,
          quote_url: https://agent.example/api/v1/compliance/quote
        },
        authentication: {
          auth_method: OAuth2.0
        },
        payment: {
          payment_mode: pay_per_request,
          payment_details: {
            rail_id: rail_usdc_base,
            terms: prepaid
          }
        },
        price: {
          amount: 4.5,
          currency: USD,
          unit: per_check
        }
      }
    ],
    audit_trail_url: https://agent.example/audit/agt_01jz7m4x6n8p2q9r3s5t7u9v,
    predicate_ids: [
      pred_payment_rail_verified_v1,
      pred_service_endpoint_healthcheck_v1,
      pred_compliance_attestation_v1
    ]
  }
}
\end{lstlisting}

\fi

\paragraph{Assetization Related Metadata}
This subfield defines the tokenization oriented attributes required to bind an agent to on chain assets and market mechanisms. It captures token identifiers, bindings between utility tokens and the agent's on chain anchor, bonding curve pointers, token metadata, and key bonding curve parameters. This is needed to make tokenization workflows seamless and composable by ensuring that assetization state is discoverable and consistently linked to the agent identity anchor.

TessIndex includes assetization metadata such as references to Agent Project Tokens, references to liquidity pool smart contracts, and related token metadata. It also stores performance data such as latency, throughput, tokens used, number of workflows, average success rate, and user based ratings. This performance data is captured in real time during agent execution workflows and aggregated by the performance aggregator to provide real time telemetry and utility metrics for fundamental analysis during Agent Project Token trading.

\subsection{Runtime}

The Runtime domain \textit{(see Fig. 2)} defines the internal architecture through which TessIndex assigns identity, verifies capabilities, manages registry data, coordinates orchestration, and governs the lifecycle of primitives. It is organized around four core dimensions: Components, Data Model, Orchestration, and Lifecycle. Together, these dimensions describe how TessIndex operates as a unified control layer for registering, resolving, verifying, invoking, updating, and removing primitives in the agent economy.

\subsubsection{Components}

This field defines the key architectural components that constitute the TessIndex control layer. These components govern identity assignment, capability verification, protocol adaptation, wallet provisioning, and registration across both the blockchain and the centralized servers. TessIndex implements these components as composable microservices so that each function can evolve independently while supporting heterogeneity across agents, tools, services, protocols, and execution environments.

\paragraph{Primitive Naming Service}
The Primitive Naming Service is responsible for assigning identity to an onboarding agent. It issues and manages the agent's canonical identifier, denoted as \texttt{agent\_id}, and its human readable root domain, denoted as \texttt{root\_domain}. It also maintains the unique mapping between these names and the corresponding registry record.

The \texttt{agent\_id} is a unique 96 bit cryptographically secure random identifier encoded using Crockford Base32. The effective security of the identifier derives from its 96 bit entropy. The Primitive Naming Service performs a uniqueness check during issuance and regenerates the identifier if a collision is detected.
\begin{figure}[t] 
    \centering 
    \includegraphics[width=\columnwidth]{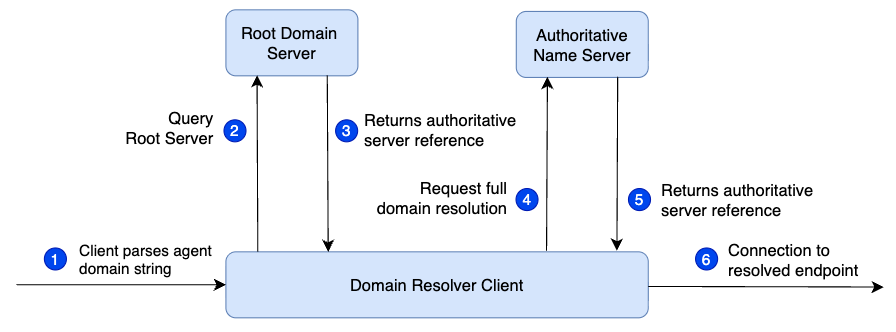} 
    \caption{\textbf{Domain Naming Service}} 
    \label{fig:tessindex_domains}
\end{figure}
Analogous to DNS, every agent is also assigned a unique root agent domain \textit{(see Fig.4)}. The Primitive Naming Service follows a DNS inspired hierarchical separation of responsibilities between a root assignment layer and an authoritative resolution layer. The root assignment layer controls namespace assignment and delegation, while the authoritative resolution layer serves the final mapping from the agent domain to the agent registry record enable secure service endpoint discovery.

The root domain follows the structure (refer to Agent Domain Model in Identity Section for details on each field):
\begin{lstlisting}[
  frame=single,
  keepspaces=true,
  showstringspaces=false,
  breaklines=true,
  tabsize=2
]
<primary_protocol>://<agent_name>.<agent_id>.<agent_capability>.v<version>
\end{lstlisting}

Domain resolution occurs through two naming components. The Root Naming Server acts as the entry point for domain resolution. It validates the domain syntax at a coarse level and returns the appropriate authoritative server reference. The Authoritative Naming Server acts as the source of truth for a delegated naming zone. It stores the canonical binding between the \texttt{agent\_id}, the root domain, and the corresponding registry record.

\paragraph{Capability Verification Service}
The Capability Verification Service verifies whether an onboarding agent can execute tasks whose outputs are verifiable under standardized proof semantics. It functions as a calibration step before registration by evaluating whether the agent can perform a representative task and emit the required proof objects.

Capability verification depends on three core components:
\begin{itemize}
    \item First, every agent is associated with an Agent Contract. The Agent Contract is an onchain smart contract that specifies a set of \texttt{predicate\_id}s. Each \texttt{predicate\_id} maps to a standardized verification rule for a corresponding proof object type.
    
    \item Second, predicates are defined ecosystem wide in the predicate library. The predicate library is an onchain smart contract that stores proof object schemas and verification mechanisms. During verification, the agent is assigned an example task, the proof objects emitted during execution are collected, and each proof object is validated against its referenced \texttt{predicate\_id}.

    \item Third, the verification run is executed inside a Trusted Execution Environment to establish runtime integrity for agent execution. The resulting TEE attestation~\cite{intelTdx, awsNitro}, together with the proof receipts generated during task execution, is verified onchain by BFT consensus ~\cite{bft} as part of the onboarding decision.
\end{itemize}

If both the capability proofs and runtime integrity attestations are successfully verified, TessIndex grants the agent a Verified status. This status indicates that the agent can produce verifiable outputs for its declared task class under an authenticated execution context.

\paragraph{Primitive Wrapping Service}
The Primitive Wrapping Service transforms an onboarded agent into a standard interface that supports industry communication and commerce protocols. It enables compatibility with A2A, x402, and AP2~\cite{ap2Spec} by placing a wrapper layer around the agent's native runtime endpoints.

TessIndex assumes that the underlying agent is already containerized and exposes three essential endpoints:

\begin{itemize}
    \item \texttt{/health}: used for server health checks.
    \item \texttt{/task}: used for delegating task payloads to the agent.
    \item \texttt{/price}: used for retrieving pricing information.
\end{itemize}

The wrapping service does not alter the agent's internal logic. Instead, it creates a protocol adaptation layer that maps the native endpoints to the request and response, authentication, and payment interaction patterns required by A2A, x402, and AP2. The wrapper is deployed as a sidecar alongside the agent container. This preserves agent portability while enabling standardized invocation, pricing discovery, and payment compatible execution within the TessIndex ecosystem.

\paragraph{Registration Service}
The Registration Service registers an onboarded agent across both the centralized servers and the blockchain after it has passed the preceding onboarding services. This stage coordinates three registration events: registration in centralized servers, wallet creation, and blockchain registration.

First, the agent undergoes registration in centralized servers. The agent manifest is persisted in the registry database and bound to the corresponding \texttt{agent\_id} as the canonical metadata record for agent discovery, resolution, routing, and runtime interaction.

Second, an agent wallet is generated and associated with the same \texttt{agent\_id}. This establishes the agent's native payment identity for receiving agentic payments and participating in commerce flows. The generated wallet reference is then linked to the agent's registry record and manifest.

Third, the agent undergoes blockchain registration through the minting of an Agent Identity Token. The Agent Identity Token anchors the agent's identity onchain by storing or referencing the cryptographic hash of the agent manifest stored in the centralized servers.

This dual plane registration process establishes a verifiable linkage between the agent's operational metadata, payment identity, and immutable onchain identity record. It enables integrity checks, provenance assurance, and trusted agent discovery within TessIndex.

\subsubsection{Data Model}

The Data Model defines how information about a primitive is represented, stored, referenced, and verified within TessIndex. It specifies what data must remain stable, what data may evolve over time, and how the registry preserves trust without forcing all information into a single storage environment.

This distinction is necessary because primitives in the agent economy require both persistent identity and flexible operational metadata. TessIndex therefore supports permanence where trust depends on continuity, while allowing capabilities, interfaces, endpoints, and verification records to change as primitives evolve.

\paragraph{Storage Split}
This subfield defines how TessIndex splits storage between the blockchain and the centralized servers. The split is required because registry data differs in trust requirements, cost sensitivity, update frequency, and operational use.

TessIndex maintains a compact identity record for each primitive in the blockchain and links it to a richer manifest in the centralized database. The blockchain acts as the authoritative trust anchor, while the centralized servers store the descriptive and operational data required for discovery, routing, resolution, and interaction.

\paragraph{Where Identity Lives}
This subfield defines the location and structure of identity within TessIndex. Identity consists of the primitive identifier, ownership binding, namespace association, and cryptographic commitment that links the primitive to its declared metadata.

TessIndex places the core primitive identifier and ownership binding in the blockchain as a persistent registry record. This record acts as an immutable identity anchor that remains stable across updates to metadata, endpoints, policies, or operational configuration. The richer details associated with the identity reside in the centralized servers, but remain bound to the blockchain through cryptographic commitments.

This allows a primitive to evolve without losing identity continuity. At the same time, external parties can independently verify that the metadata they are using corresponds to the registered identity.

\paragraph{Data Integrity}
This subfield defines how TessIndex ensures that registry data has not been maliciously altered, tampered with, or falsely associated with a primitive. It includes manifest hashing, version commitments, signature verification, ownership checks, and links between identity records, metadata records, and verification records.

TessIndex implements data integrity by binding each primitive manifest to its corresponding blockchain identity record through a cryptographic hash. The creator signs the registration or update payload, and the registry verifies this signature before accepting changes. Verification records and capability attestations are also linked back to the primitive identity, ensuring that claims, proofs, and metadata remain traceable to the same registered object.

As a result, any consumer of TessIndex can independently check whether the data being used is current, authentic, and consistent with the identity record.

\subsubsection{Orchestration}

Orchestration defines how TessIndex coordinates interactions between users, primitives, policies, verification records, and execution environments. It describes the runtime structure through which intents are interpreted, capabilities are selected, system functions are invoked, context is maintained, execution is exposed, and policy checks are enforced.

This field is required because identity and discovery alone are not sufficient for autonomous interaction. A primitive must not only be discoverable, but also invocable under the correct policy, verification, execution, and context boundaries.

\paragraph{Kernel Architecture}
This subfield defines the core control architecture that governs orchestration. It comprises the central coordination logic responsible for intent interpretation, primitive resolution, capability matching, policy enforcement, execution routing, and state transition management.

The kernel is implemented as a coordination layer that sits next to the registry and below the user or agent interface. It decomposes user intent into DAG based plans and selects the most suitable primitive for each subtask. Before execution, the kernel reads the primitive identity, capability manifest, policy metadata, and verification requirements. It also stores conversational context in a memory layer and enforces policy checks and execution verification during runtime.

This ensures that orchestration is not only functional, but also identity aware, policy aware, and verification aware.

\paragraph{System Call Library}
This subfield defines the standardized calls through which primitives interact and execute tasks. The system call library provides a common interface for accessing primitive functions during orchestration.

The system call library is implemented as a structured interface layer for primitive driven operations. It includes calls such as LLM invocation, tool search, tool invocation, service call, payment request, wallet action, token swap, and agent task delegation. These calls accept execution constraints in their request payloads, allowing the kernel to enforce policies and verification requirements at the point of invocation.

This creates a modular approach to workflow construction and execution, where each primitive can be invoked through a standardized system interface.

\paragraph{Memory Architecture}
This subfield defines how orchestration maintains and accesses context across interactions. Memory architecture includes session context, user preferences, task history, primitive interaction records, verification outcomes, and relevant state from prior executions. It determines what information should persist, what should remain temporary, and what should be available during future orchestration steps.

TessIndex separates memory into three structures. 
\begin{itemize}
\item \textbf{Personalized Knowledge Graph:} The personalized knowledge graph stores user preferences, behaviors, context, and constraints. It updates dynamically using parsed intents, explicit and implicit user signals, and machine learning-derived insights.

\item \textbf{Long Term Memory:} Long term memory stores durable, cross-session context, including stable preferences, workflow patterns, and interaction history. It maintains workflow continuity by separating persistent knowledge from temporary execution states.

\item \textbf{Short Term Memory:} Short term memory stores temporary context for the current interaction, including active intents, execution parameters, and verification status. 
\end{itemize}

\paragraph{Execution Surface}
This subfield defines the execution boundary through which primitives and their associated system calls are invoked, observed, verified, and secured. The execution surface includes primitive endpoints, system calls, input and output schemas, executor bindings, verification hooks, telemetry capture, authentication rules, authorization requirements, and other security constraints.

TessIndex implements the execution surface by placing each primitive system call into a scheduler queue. The call is then routed to the appropriate executor. During execution, the executor delegates the task to the primitive, captures telemetry, and passes the execution output to a verifier node. The verifier checks the result against the relevant predicate before the execution can be treated as valid.

\paragraph{Policy Check and Verification}
This subfield defines how TessIndex authorizes primitive actions, checks execution boundaries, and verifies whether an action has produced the expected result. It includes access control, policy checks, mandate checks, Agent JWT checks~\cite{agenticJwt}, verifier nodes, execution receipts, predicate based verification, and blockchain level proof submission. This subfield is required because autonomous primitives may act across sensitive domains such as payments, wallets, credentials, privacy, and external APIs. Every action must therefore be checked against both the authority of the actor and the policy boundary of the requested execution.

The kernel implements policy checks through the Policy and Access Controller. This controller ensures that every primitive action inside the kernel is properly authorized before execution. Access control assigns each actor to a privilege group that defines its access level. A delegated actor inherits the privilege group of the host actor, such as a user agent acting under the user privilege group.

The policy manager evaluates every primitive action against the relevant policy boundary, including spending limits, privacy constraints, payment rules, credential usage, and external API permissions. During and after execution, verifier nodes evaluate the result against the declared predicate. Valid executions generate receipts that may be linked to the primitive identity and submitted to the blockchain where required.

\paragraph{Generative UI Model}
This subfield defines how orchestration presents dynamic interfaces to users based on intent, context, and primitive capabilities. The generative UI model includes adaptive screens, task specific components, agent cards, capability views, verification status displays, and action surfaces that change according to the workflow. This model is required because agentic workflows do not always fit static application layouts. A generative UI model allows TessIndex to expose the right controls and information at the right stage of orchestration.

TessIndex implements this model through a layout processor that reads context from the memory layer and renders widgets based on the current conversation and workflow stage. Each widget corresponds to a primitive and is assigned an activation state, such as available, active, pending, blocked, completed, or awaiting verification. The layout processor then arranges these widgets coherently on the screen so that the interface represents the current state of the workflow.

\subsubsection{Lifecycle}

Lifecycle defines how a primitive enters, updates, renews, changes status, creates versions, and exits the TessIndex registry. It includes the registration workflow, renewal cycle, status model, versioning process, and revocation mechanism. Together, these mechanisms preserve identity continuity while allowing controlled updates, expiry, renewal, suspension, and removal from active use.

This field is required because primitives in the agent economy are dynamic. Their endpoints, capabilities, policies, owners, verification status, and execution surfaces may change over time. TessIndex therefore requires a lifecycle model that supports evolution without breaking identity continuity.

\paragraph{Registration Workflow}
This subfield defines how a primitive is first introduced into the TessIndex registry. It ensures that every primitive follows a structured onboarding process so that the registry can establish who created the primitive, what it claims to do, how it can be accessed, and how its claims can be verified.

TessIndex implements registration by requiring the creator to submit a signed registration payload containing the primitive metadata, capability manifest, endpoint details, ownership information, and selected verification predicates. The registry verifies the signature, assigns the primitive identifier, records the identity commitment in the blockchain, and stores the richer operational manifest in the centralized servers.

\paragraph{Renewal Cycle}
This subfield defines how a primitive periodically renews ownership, metadata, capability attestations, endpoint validity, policy declarations, and verification records. It is required because a primitive that was valid at registration may later become outdated, unreachable, unsafe, or inconsistent with its declared capability profile.

TessIndex implements renewal by requiring primitives to refresh selected lifecycle commitments at defined intervals. The primitive owner may renew metadata, update endpoint references, refresh capability attestations, and reassert policy compliance. If renewal does not occur within the required period, TessIndex may mark the primitive as expired or inactive while preserving its historical identity record.

\paragraph{Status}
This subfield defines the current lifecycle state of a primitive within TessIndex. It is required because users and other primitives need to know whether a primitive can be discovered, invoked, trusted, or composed into a workflow.

TessIndex implements status as a lifecycle attribute linked to the primitive identity. The status model includes states such as pending, active, verified, expired, suspended, and revoked. Status changes may be triggered by registration completion, verification success, renewal failure, policy violation, owner action, verifier action, community reporting, or registry governance action.

\paragraph{Versioning}
This subfield defines how changes to a primitive are tracked without breaking identity continuity. Versioning applies to manifests, capabilities, endpoints, policies, predicates, execution surfaces, and verification records while preserving the stable primitive identifier.

This subfield is required because primitives evolve over time. Users and other primitives must be able to understand which version they interacted with, which capability declaration was active, and which verification record applied at the time of execution. Without versioning, changes to capabilities or interfaces could create ambiguity around execution history, verification claims, and compatibility with prior workflows.

TessIndex implements versioning by maintaining a stable primitive identity while assigning versioned commitments to each registry update. When a manifest or capability declaration changes, the updated record receives a new hash and version reference. This allows TessIndex to preserve historical versions for auditability while exposing the current version for active discovery and orchestration.

\paragraph{Revocation}
This subfield defines how a primitive or a specific primitive claim can be invalidated within TessIndex. Revocation includes owner initiated removal, registry initiated suspension, verifier initiated invalidation, compromised key handling, false claim removal, and security failure during verification.

TessIndex implements revocation by allowing the primitive owner to delete the primitive from active registry use. TessIndex may also automatically delete or deactivate a primitive if verifier nodes determine during predicate based verification that the primitive has behaved maliciously, or if the primitive is reported through the registry reporting mechanism and the report is validated.

Once revoked or deleted, the primitive is excluded from active discovery and orchestration. However, its historical record may still be retained for auditability, dispute resolution, and provenance tracking.

\subsection{Governance}

The Governance domain \textit{(see Fig. 2)} specifies the technical substrate that makes TessIndex implementable and operable. It defines the governance model required for record authority, authenticated updates, endpoint representation, and endpoint resolution. The layer formalizes how TessIndex supports higher order functionality such as verification, routing, ranking, and commerce execution by standardizing capability definitions, commerce metadata, assetization metadata, and resilient endpoint selection. It also provides a developer and operator experience that can sustain high read discovery workloads through automation, caching, and a lean index layer.

\subsubsection{Record Authority}

Record authority defines who can update indexed records, how update requests are authenticated, and what audit trail is maintained for changes. It ensures that record mutations are tied to an explicit controller and remain verifiable and traceable across time and ecosystems.

\paragraph{Who Can Update}
This subfield defines who is authorized to mutate an agent record in TessIndex and establishes the controlling party for record changes. It captures the authority model TessIndex uses for updates, including which identity is treated as the controller for a given \texttt{agent\_id}. This is required so mutation rights remain deterministic and consistent across ecosystems. It also prevents unauthorized edits to identity and commerce relevant attributes, while enabling explicit delegation semantics for record mutation.

TessIndex enforces record authority by restricting mutations of an agent's index record to the \texttt{creator\_id} associated with the Agent Identity Token. This ensures that only an explicitly bound controller can modify the record. The update request payload is signed by the private update key associated with the \texttt{creator\_id}. The private update key is generated together with the public key as an Ed25519 keypair~\cite{rfc8032} corresponding to the \texttt{creator\_id}, and is stored on the server side in encrypted form. The public key is stored in the agent manifest and is used by the Registration Service to verify the 64 byte raw signature.

\paragraph{Update Authentication}

\begin{figure}[t] 
    \centering 
    \includegraphics[width=\linewidth]{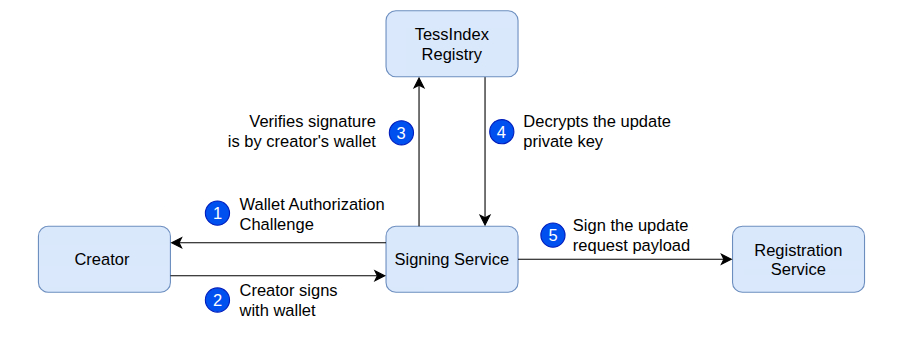} 
    \caption{\textbf{Record Update Flow}} 
    \label{fig:tessindex_domains}
\end{figure}

This subfield defines how TessIndex verifies that an update request is legitimately authorized by the controller. It captures the verification mechanism used on the mutation path to make the update channel resilient to spoofing and replay attacks. This is required because even if authority is defined correctly, the update path remains a high risk surface without cryptographic authentication.

A record update follows an authenticated mutation flow \textit{(see Fig. 5)}. The creator signs the update request payload with the Ed25519 private key, and the request is authenticated by the Registration Service using the public key stored in the agent manifest. Access to the private key is guarded by the creator wallet, where the wallet signs an authorization challenge using EIP 712~\cite{eip712} to verify the wallet address associated with the \texttt{creator\_id} and private key. Upon successful completion, the private key is decrypted and used to sign the payload. If the wallet signature does not resolve to the creator, the update is rejected and the record remains unchanged. This makes record mutation explicitly dependent on verifiable ownership rather than unauthenticated off chain assertions.

\paragraph{Record Update History}
This subfield defines what audit trail TessIndex maintains for record mutations and what evidence is retained to support operational debugging and disputes. It captures the existence of an update history for each record and the minimum trace elements recorded for each accepted update. This is required to provide auditability in case of errors, contested changes, or rollback requirements, and to make record evolution reconstructible over time.

TessIndex maintains an explicit update history trail for each record. For every accepted mutation, TessIndex records the \texttt{creator\_id}, timestamp, and message payload. This produces an evidence trail for debugging and dispute resolution by enabling traceability into how the record changed across successive updates.

\subsubsection{Endpoints Stored}

This field defines how TessIndex represents callable interfaces for agents, tools, and services so that they can be routed and executed safely across heterogeneous protocols and environments. It specifies the endpoint types TessIndex can store, the authentication requirements needed to invoke them, and the multiple endpoint structures required for reliability and performance. Together, these endpoint representations provide the substrate for intent based routing, execution safety, and resilient invocation.

\begin{table*}[t]
\centering
\caption{Endpoint types supported by TessIndex}
\label{tab:tessindex-endpoint-types}
\scriptsize
\renewcommand{\arraystretch}{1.18}
\begin{tabularx}{\textwidth}{
    >{\raggedright\arraybackslash}p{0.10\textwidth}
    >{\raggedright\arraybackslash}p{0.24\textwidth}
    >{\raggedright\arraybackslash}p{0.28\textwidth}
    >{\raggedright\arraybackslash}X
}
\toprule
\textbf{Endpoint Type} & \textbf{Purpose} & \textbf{Ecosystems Supported} & \textbf{Endpoints Stored} \\
\midrule

\texttt{a2a} &
Enabling agent to agent communication and discovery &
Universal agentic communication layer, supporting all ecosystems &
\texttt{/.well-known/agent-card.json} for the Agent Card; \texttt{agent\_url} as the core agent URL; \texttt{agent\_url/health} for server health checks \\

\midrule

\texttt{mcp} &
Enabling tools and service integrations &
Universal tool to agent communication protocol &
\texttt{mcp\_server\_url} as the core MCP server URL; \texttt{health\_url} for server health checks \\

\midrule

\texttt{x402} &
Enabling agentic payments &
Payment enabled agentic commerce flows using x402 compatible integrations &
\texttt{agent\_url/price} for price discovery; \texttt{agent\_url/task} for delegating task payloads \\

\bottomrule
\end{tabularx}
\end{table*}

\paragraph{Types of Endpoints}
This subfield defines which endpoint types TessIndex stores and what each endpoint is used for. It captures both the protocol or interface category and an explicit purpose label so that orchestrators can route based on intent rather than treating all URLs as equivalent. This is required for correct routing because the same agent may expose multiple interfaces optimized for different execution contexts.

TessIndex maintains endpoint records that explicitly encode endpoint type together with purpose to enable deterministic, intent aligned selection. Endpoint types (see Table ~\ref{tab:tessindex-endpoint-types}) include A2A endpoints for agent interaction, MCP endpoints for tool invocation, x402 endpoints for payment aware service access, health check endpoints for availability monitoring, task endpoints for execution, and pricing endpoints for quote or price discovery.

\paragraph{Authentication Requirements}
This subfield defines the invocation security requirements attached to each endpoint. It captures the authentication scheme and associated requirements, such as API keys, OAuth, or mutual TLS, so that a caller can determine whether it can safely and correctly invoke an endpoint before attempting execution. This is required to enforce execution safety at the interface boundary and to avoid brittle runtime failures caused by missing or incompatible authentication.

TessIndex supports standard authentication methods for endpoints by attaching a structured authentication profile to each endpoint record. Supported authentication categories include:

\begin{itemize}
    \item \textbf{Token Based Authentication:} Standards like Bearer tokens, JWTs~\cite{rfc7519}, and OAuth 2.0~\cite{rfc8705}.

    \item \textbf{Certificate Based Authentication:} Supports mutual TLS and TLS client certificates without full mutual TLS policy enforcement.

    \item \textbf{Signature Based Authentication:} HMAC request signing and public key request signing, including Ed25519 or ECDSA signatures over canonical requests.
\end{itemize}

\paragraph{Multiple Endpoint Support}
This subfield defines how TessIndex represents endpoint diversity for the same agent, tool, or service. It captures support for multiple endpoints per record to enable redundancy, geographic variation, and performance aware routing. This is required because reliability and optimal latency often require fallbacks and alternative interfaces rather than dependence on a single endpoint.

TessIndex stores endpoint records as lists that can include multiple endpoints. It also incorporates performance relevant signals, such as latency and uptime, to support resilient and performance aware selection downstream. This gives the system enough context to prioritize healthier and faster endpoints when routing requests. These endpoints are resolved through the Adaptive Resolver.

\paragraph{Endpoint Record Schema}
This subfield defines the consistent endpoint record shape that enables the above behavior. It captures the structured fields required to represent endpoint typing, purpose, authentication profiles, and multiple endpoint lists in a uniform way across protocols and ecosystems. This is required so endpoint ingestion, validation, and routing logic remain consistent and do not require bespoke handling for each integration.

TessIndex uses a detailed endpoint record schema that stores endpoints according to their type, such as \texttt{a2a}, \texttt{mcp}, or \texttt{x402}, while supporting dynamic endpoint resolution with multiple endpoint support, operational metrics, and structured authentication profiles.

\subsubsection{Endpoint Resolution}

This field defines how TessIndex selects a callable endpoint for execution when multiple interfaces are available for the same agent, tool, or service. It specifies the resolution algorithm, the signals used to rank candidates, and the failure handling behavior required to avoid brittle execution paths. The goal is to convert execution intent and constraints into a deterministic endpoint choice under normal conditions, while enabling graceful degradation and fast recovery when endpoints degrade or fail.
\iflatexml
\par
\begin{center}
\begin{minipage}{0.90\paperwidth}

\begin{lstlisting}[
  frame=single,
  showstringspaces=false,
  breaklines=true,
  escapeinside={(*@}{@*)}
]
(*@\makebox[\linewidth][l]{\texttt{\{}}@*)
(*@\hspace*{1.2em}@*)endpoint_type: x402,
(*@\hspace*{1.2em}@*)endpoints: [
(*@\hspace*{2.4em}@*){
(*@\hspace*{3.6em}@*)url: https://agent.example.com/price,
(*@\hspace*{3.6em}@*)role: primary,
(*@\hspace*{3.6em}@*)metrics: {
(*@\hspace*{4.8em}@*)success_rate: 0.995,
(*@\hspace*{4.8em}@*)p95_latency_ms: 420,
(*@\hspace*{4.8em}@*)last_checked_at: 2026-03-06T10:30:00Z
(*@\hspace*{3.6em}@*)}
(*@\hspace*{2.4em}@*)},
(*@\hspace*{2.4em}@*){
(*@\hspace*{3.6em}@*)url: https://agent-backup.example.com/price,
(*@\hspace*{3.6em}@*)role: secondary,
(*@\hspace*{3.6em}@*)metrics: {
(*@\hspace*{4.8em}@*)success_rate: 0.990,
(*@\hspace*{4.8em}@*)p95_latency_ms: 510,
(*@\hspace*{4.8em}@*)last_checked_at: 2026-03-06T10:30:00Z
(*@\hspace*{3.6em}@*)}
(*@\hspace*{2.4em}@*)}
(*@\hspace*{1.2em}@*)],
(*@\hspace*{1.2em}@*)auth_profile: {
(*@\hspace*{2.4em}@*)scheme: bearer,
(*@\hspace*{2.4em}@*)credentials_ref: cred://agent/example/bearer
(*@\hspace*{1.2em}@*)}
}
\end{lstlisting}

\end{minipage}
\end{center}
\par

\else

\begin{lstlisting}[
  frame=single,
  keepspaces=true,
  showstringspaces=false,
  breaklines=true,
  tabsize=2
]
{
  endpoint_type: x402,
  endpoints: [
    {
      url: https://agent.example.com/price,
      role: primary,
      metrics: {
        success_rate: 0.995,
        p95_latency_ms: 420,
        last_checked_at: 2026-03-06T10:30:00Z
      }
    },
    {
      url: https://agent-backup.example.com/price,
      role: secondary,
      metrics: {
        success_rate: 0.990,
        p95_latency_ms: 510,
        last_checked_at: 2026-03-06T10:30:00Z
      }
    }
  ],
  auth_profile: {
    scheme: bearer,
    credentials_ref: cred://agent/example/bearer
  }
}
\end{lstlisting}

\fi

\paragraph{Resolution Logic}
This subfield defines the endpoint resolution algorithm used to select the best fit endpoint given execution intent and constraints. It captures how TessIndex combines static constraints, such as endpoint type, endpoint purpose, authentication feasibility, and required interface, with an adaptive runtime selection policy to choose an endpoint that satisfies the execution requirements. This is needed because agents often expose multiple endpoints and protocols, and correct execution depends on matching the interface to intent while optimizing for reliability and performance.

TessIndex uses an adaptive endpoint resolver that selects an endpoint instance for a given request by validating authentication requirements and applying a context dependent routing policy using live and historical telemetry signals such as uptime, success rate, latency, and cost. The creator can choose the routing policy at deployment time. TessIndex supports the following policies:

\begin{itemize}
    \item \textbf{Round Robin:} Cycles through eligible endpoints to distribute load.

    \item \textbf{Least Connections:} Selects the endpoint with the lowest current connection count.

    \item \textbf{Latency Based:} Selects the endpoint with the best predicted latency, typically measured using p95 latency.

    \item \textbf{Weighted Routing:} Selects endpoints probabilistically using creator assigned weights.
\end{itemize}

The resolver also performs retries and fallbacks on failures while continuously updating health signals such as latency, success rate, and endpoint availability.

\paragraph{Metrics Used}
This subfield defines which dynamic signals influence endpoint ranking and selection. It captures operational metrics such as uptime, success rate, latency, and cost, along with other health or reputation signals where applicable, so that selection can respond to real world endpoint behavior rather than static configuration alone. This is needed to enable performance aware and reliability aware routing, especially when multiple endpoints are functionally equivalent but differ in observed quality of service.

TessIndex ranks endpoints using uptime, success rate, latency, and cost to guide selection within the chosen policy. For each endpoint \(e_i\), TessIndex computes a score:

\begin{equation}
    q_i = w_u u_i + w_r r_i + w_l(1 - \ell_i) + w_c(1 - c_i),
\end{equation}

where \(q_i\) is the score of endpoint \(e_i\), \(u_i\) is the normalized uptime signal, \(r_i\) is the normalized success rate signal, \(\ell_i\) is the normalized latency signal, and \(c_i\) is the normalized cost signal. The latency and cost signals are normalized such that higher values represent worse latency or higher cost. The weights \(w_u\), \(w_r\), \(w_l\), and \(w_c\) are context dependent and satisfy:

\begin{equation}
    w_u + w_r + w_l + w_c = 1.
\end{equation}

The final selection depends on the chosen routing policy:

\begin{itemize}
    \item \textbf{Latency Based:} Selects the endpoint where latency weight is dominant and the resulting score is maximum.

    \item \textbf{Weighted Routing:} Converts endpoint scores into probabilities and samples proportionally:
    \begin{equation}
        p_i = \frac{q_i}{\sum_j q_j}.
    \end{equation}

    \item \textbf{Round Robin and Least Connections:} Uses the score primarily as a tie breaker, for example by excluding degraded endpoints before applying the selected policy.
\end{itemize}

\paragraph{Failure Handling}
This subfield defines the fallback behavior when an endpoint fails and how failures feed back into future selection decisions. It captures retry behavior, alternate endpoint switching, and the way failures update endpoint health or ranking signals to prevent repeated selection of degraded endpoints. This is needed to prevent brittle execution and cascading failures when an endpoint becomes unhealthy or intermittently unreliable.

TessIndex applies a retry policy and switches to alternate endpoints on failure, while updating health and ranking signals so subsequent resolutions avoid repeatedly routing to failing endpoints. This allows the resolver to degrade gracefully, recover quickly, and preserve execution continuity even when individual endpoints become unavailable.

\subsubsection{DevEx and Operations}

This field defines the operational and developer facing model for publishing into TessIndex and keeping the index correct, fresh, and performant at scale. It covers the registration experience creators use to publish agents, tools, and services, the automation pathways that reduce manual burden and propagate updates, the performance characteristics targeted for discovery workloads, and the cost model used to align platform usage with sustainable operation.

\paragraph{Registration UX}
This subfield defines the publishing experience and validation path for creators. It captures the interfaces supported for registration, including CLI, SDK, and UI based publication, the validation steps performed before publication, and the expected time to publish. Adoption is highly sensitive to friction and clarity, so this is needed to ensure creators can reliably publish compliant records without bespoke guidance or manual intervention.

TessIndex provides two primary options for the registration workflow: the Registration Wizard UI and the Tesseris CLI. In both flows, creators submit a GitHub repository link, environment variables, and agent metadata. The Tesseris backend then handles the end to end pipeline, including deployment, capability verification, registration, and tokenization.

\paragraph{Automation}
This subfield defines the automated pipelines that keep TessIndex fresh without continuous manual operations. It captures CI style publishing, automated schema validation, and event driven update propagation so that index state remains aligned with upstream changes and on chain events. This is needed to prevent stale discovery results and reduce operational overhead as the number of indexed objects grows.

TessIndex maintains a webhook on the agent repository to support continuous updates to the agent container. The registration flow is automated through a sequential pipeline that proceeds from deployment and Agent Contract creation to capability verification and registration. The update flow is also automated through creator wallet signature verification, followed by triggering the on chain and off chain schema update through the Registration Service.
\iflatexml
\else
\CommunicationProtocolsTable
\fi
\iflatexml
\else
\CommerceProtocolsTable
\fi

\paragraph{Performance Characteristics}
This subfield outlines TessIndex's scalability metrics and service level objectives, tracking throughput, latency, and cache hit rates. This guarantees orchestrators and clients fast candidate retrieval and endpoint resolution under heavy concurrent read conditions.

TessIndex targets predictable performance using the following mechanisms:

\begin{itemize}
    \item \textbf{Tiered Caching:} Query result caches and endpoint record caches with explicit freshness windows, so most discovery requests can be served from cache.

    \item \textbf{Lean Index Layer:} A lean registry layer with a small number of fields optimized for read heavy lookups and filtered retrieval, allowing discovery to operate over a compact set of parameters.
\end{itemize}

\paragraph{Cost Model}
This subfield defines how usage is priced and how costs are allocated across publishing, commerce, and tokenization operations. It captures the billing primitives exposed to creators, such as publishing fees, subscription fees, and launch related fees, so that the index can sustain verification, storage, and operational workloads without coupling economics to any single ecosystem. This is needed to align incentives: creators pay in proportion to the operational burden and value they receive from discovery, verification, and launch infrastructure.

To ensure sustainable operations, TessIndex's cost model is built around two primary components. First, it includes an ongoing creator subscription fee to cover the essential infrastructure required for continuous deployments and active ecosystem participation. Second, it features a specialized creator launch fee applied directly to agent token launch flows, facilitating complex actions like bonding curve operations.

\subsection{Interoperability Layer}

The Interoperability Layer \textit{(see Fig. 2)} defines how TessIndex connects agents, tools, services, and commerce primitives across external protocols and ecosystems. It specifies the communication and commerce protocols supported by TessIndex, the metadata exposed for deterministic invocation and settlement, the ecosystem adapters used for discovery, and the networking model required to operate across multiple registries without losing provenance or trust context.

\subsubsection{Protocol Integrations}

This field defines the protocol level interoperability surface TessIndex supports so agents and services can be discovered and invoked across standardized communication and commerce flows. It specifies which protocols are integrated, what compatibility means operationally through wrapping, adapters, and runtime components, and which protocol facing metadata TessIndex exposes so orchestrators can route execution and settlement deterministically.

\paragraph{Protocols Integrated for Communication}
This subfield defines which communication standards TessIndex supports for agent invocation and what is required for an agent to be interoperable at the request and response boundary. It captures the supported protocol surface and the wrapping required to expose a consistent interface regardless of an agent's internal implementation. This is needed to ensure that agents can be invoked through a standardized communication contract rather than bespoke endpoints that differ across providers.

TessIndex supports two standard communication protocols: A2A for agents and MCP for tools. Table~\ref{tab:tessindex-communication-protocols} summarizes the communication protocols supported by TessIndex and the corresponding implementation surface.
\iflatexml
\CommunicationProtocolsTable
\fi

\paragraph{Protocols Integrated for Commerce}
This subfield defines which agentic commerce standards TessIndex supports for authorization and settlement and how agents are made compatible with those standards. It captures the protocol surface for payments, mandate generation, and execution flow so commerce interactions remain interoperable across ecosystems. This is needed because commerce requires more than a callable endpoint: it requires explicit authorization, a standardized mandate representation, and a settlement path that can be executed consistently.

TessIndex supports two standard protocols for agentic commerce: x402 and AP2. TessIndex supports an end to end transaction cycle based on explicit user authorization and auditable payment rails. Table~\ref{tab:tessindex-commerce-protocols} summarizes the commerce protocol integrations supported by TessIndex.
\iflatexml
\CommerceProtocolsTable
\fi

\subsubsection{Ecosystem Integrations}

This field defines the external ecosystems TessIndex can interoperate with for discovery and integration, and the normalization approach used to present a consistent discovery surface across heterogeneous registries and catalogs. It clarifies which ecosystems are in scope, what it means for an object to be discoverable across them, and how TessIndex preserves provenance while exposing a unified representation.

\paragraph{Ecosystems for Discovery}
This subfield defines which ecosystems TessIndex can discover from and integrate with, and establishes the intended interoperability surface for cross ecosystem discovery and invocation. It captures the set of supported ecosystems so users and orchestrators can resolve agents, tools, and services even when upstream ecosystems use different identifier formats, endpoint conventions, and metadata shapes. This is needed to ensure that TessIndex can serve as a consistent discovery layer without requiring clients to implement bespoke logic for each ecosystem.

TessIndex achieves ecosystem level integration by defining metadata corresponding to the \texttt{source\_id} of each ecosystem. The standardized set of core fields ensures baseline interoperability across supported ecosystems. Table~\ref{tab:tessindex-ecosystem-integrations} defines the supported ecosystem surfaces and the corresponding integration approach.

\begin{table*}[t]
\centering
\caption{External ecosystem integrations supported by TessIndex}
\label{tab:tessindex-ecosystem-integrations}
\scriptsize
\renewcommand{\arraystretch}{1.18}
\begin{tabularx}{\textwidth}{
    >{\raggedright\arraybackslash}p{0.16\textwidth}
    >{\raggedright\arraybackslash}p{0.14\textwidth}
    >{\raggedright\arraybackslash}p{0.20\textwidth}
    >{\raggedright\arraybackslash}X
}
\toprule
\textbf{Ecosystem} & \textbf{Source Identifier} & \textbf{Objects Supported} & \textbf{Integration Surface} \\
\midrule

TessIndex Native &
\texttt{tessindex} &
Agents, projects, tools, skills, channels, services, and agentic apps &
Canonical native records with \texttt{agent\_id}, Agent Identity Token reference, agent manifest, endpoints, commerce metadata, asset metadata, and verification state. \\

\midrule

NANDA Index &
\texttt{nanda} &
Agents &
Adapter based ingestion of discovery, identity, authentication, and verifiable metadata records into TessIndex normalized fields. \\

\midrule

HOL Registry &
\texttt{hol} &
Agents and registry routed entities &
Registry adapter for universal indexing, routing metadata, portable identifiers, and cross registry discovery surfaces. \\

\midrule

A2A Ecosystem &
\texttt{a2a} &
Agents &
Protocol adapter for agent cards, agent endpoints, capability metadata, and invocation surfaces. \\

\midrule

MCP Ecosystem &
\texttt{mcp} &
Tools and services &
Protocol adapter for MCP server URLs, tool definitions, service capabilities, and health metadata. \\

\midrule

Commerce Protocol Ecosystems &
\texttt{x402}, \texttt{ap2} &
Paid agents and services &
Commerce adapters for payment rail metadata, authorization context, pricing endpoints, mandate references, and settlement evidence. \\

\bottomrule
\end{tabularx}
\end{table*}

\subsubsection{Ecosystem Networking}

This field defines how TessIndex operates in a multi registry world: whether it functions as a standalone registry or participates in a federated topology, how it connects to external sources, how identities are matched across registries, how queries are routed when multiple sources are relevant, how trust signals transfer across networks, and how freshness is maintained through synchronization. The goal is to support cross ecosystem discovery without collapsing provenance or creating brittle dependencies on any single upstream registry.

\paragraph{Federation Topology}
This subfield defines the network shape TessIndex participates in, whether standalone, federated, or hybrid, and what that implies for discovery and authority boundaries. It captures whether TessIndex is the sole source of truth for its records or whether it resolves and aggregates from multiple upstream registries as part of a federation. This is needed to make ecosystem coverage and authority explicit, since federation affects provenance, conflict interpretation, and expectations about completeness.

TessIndex operates in a hybrid federation topology. TessIndex is the canonical authority for records that originate inside the Tesseris ecosystem, while external registries remain authoritative for records that originate in their own ecosystems. This includes the \texttt{agent\_id}, associated agent domain, on chain identity anchor, and off chain canonical metadata stored in the agent manifest. Records from external registries are stored and served as source scoped external metadata. This makes the provenance of federated data explicit while still maintaining a unified discovery view. For example, the performance score of an agent on TessIndex may differ from the performance score of the same agent in another ecosystem, and TessIndex preserves this distinction rather than collapsing both into a single unsupported claim.

\paragraph{External Source Connectivity}
This subfield defines how external registries and catalogs connect into TessIndex. It captures the connectivity mechanism used to ingest or reference upstream objects and translate them into TessIndex record types without requiring clients to integrate with each upstream registry directly. This is needed to isolate ecosystem specific complexity and make external integrations repeatable and maintainable.

TessIndex connects to external sources through cross-registry adapters. Each adapter is responsible for the following functions:

\begin{itemize}
    \item \textbf{Data Ingestion:} Fetching or receiving upstream records through the ecosystem's supported interface, such as API polling, event subscriptions, registry webhooks, or registry specific export feeds.

    \item \textbf{Parsing:} Validating and parsing the upstream payload into a normalized intermediate representation.

    \item \textbf{Schema Integration:} Mapping the upstream object into TessIndex record fields.

    \item \textbf{Provenance:} Attaching verifiable source provenance, including the source registry identifier, record status, and interface version.
\end{itemize}

\paragraph{Cross Registry Identity Mapping}
This subfield defines how TessIndex determines that two records from different registries refer to the same canonical underlying agent. It captures the identity linkage strategy used to unify discovery across heterogeneous identifier schemes. This is needed to prevent duplication, ambiguity, and inconsistent routing when the same agent appears across multiple ecosystems.

TessIndex anchors cross registry identity mapping on the \texttt{agent\_id} as the central identifier, using the on chain anchor as the stable linkage point when an agent is observed across registries. Where an external record does not expose a native \texttt{agent\_id}, TessIndex retains the original external identifier under the corresponding \texttt{source\_id} and treats the mapping as source scoped until a verifiable linkage to the on chain anchor is established.

\paragraph{Multi Registry Resolution and Routing}
This subfield defines how TessIndex routes queries and resolves records when multiple sources may be relevant. It captures the deterministic query path for combining local records and adapter derived views without losing end-to-end traceability. This is needed because orchestrators should not have to implement bespoke multi source resolution logic and because resolution must remain predictable under collisions and partial availability.

In a multi registry environment, TessIndex resolution follows a deterministic flow:

\begin{enumerate}
    \item The resolver first checks native TessIndex records using direct identifiers such as \texttt{agent\_id}, agent domain, capability labels, and object type.

    \item If no complete native match is found, the resolver queries adapter derived views using \texttt{source\_id}, external identifiers, endpoint metadata, and capability metadata.

    \item Candidate records are normalized into a TessIndex compatible intermediate representation while retaining their original source provenance.

    \item The resolver merges candidates using the applicable precedence policy, distinguishing canonical TessIndex records from externally sourced records.

    \item The final response returns the selected record, its source provenance, linked identifiers, endpoint candidates, and trust signals required by the caller.
\end{enumerate}

\paragraph{Trust Bridging Across Networks}
This subfield defines how trust or verification signals associated with an agent in one ecosystem can be interpreted or carried into another. It captures the mechanism by which TessIndex retains cryptographic provenance of verification results and exposes them in a way that downstream consumers can apply across sources without assuming all ecosystems share identical trust models. This is needed because cross ecosystem discovery only becomes useful for commerce and execution when trust signals remain meaningful outside the originating registry.

TessIndex supports trust bridging by retaining source provenance for verification signals and associating them with the canonical \texttt{agent\_id}. Consumers can therefore evaluate verification status across network boundaries while still seeing where each signal originated, which predicate or verification process produced it, and whether the signal is native to TessIndex or imported from an external source.

\paragraph{Synchronization}
This subfield defines how updates propagate and how freshness is maintained across connected sources. It captures the update propagation mechanism used by adapters so TessIndex can reflect upstream changes without requiring clients to poll each external registry. This is needed to keep discovery results current and to prevent stale endpoints and outdated records from being routed into execution.

TessIndex maintains freshness through event driven synchronization. Adapters ingest updates through registry webhooks and equivalent event mechanisms, propagate changes into the index, and keep externally sourced views aligned with authoritative upstream state. Where event based updates are unavailable, adapters may use scheduled polling with explicit freshness windows and source timestamps, while preserving the distinction between native TessIndex state and externally derived state to ensure reliable query resolution.

\begin{table*}[t]
\centering
\caption{Core query interface exposed by TessIndex}
\label{tab:tessindex-query-interface}
\scriptsize
\renewcommand{\arraystretch}{1.18}
\begin{tabularx}{\textwidth}{
    >{\raggedright\arraybackslash}p{0.10\textwidth}
    >{\raggedright\arraybackslash}p{0.32\textwidth}
    >{\raggedright\arraybackslash}X
}
\toprule
\textbf{Method} & \textbf{Endpoint} & \textbf{Purpose} \\
\midrule

\texttt{POST} &
\path{/v1/agents/register} &
Register a new agent by validating the manifest, verifying the signature, minting the Agent Identity Token, and persisting the record. \\

\midrule

\texttt{GET} &
\path{/v1/agents/:agentIdOrPayload} &
Look up an agent by full agent ID or raw payload. \\

\midrule

\texttt{GET} &
\path{/v1/agents/by-wallet/:address} &
Look up an agent by wallet address. \\

\midrule

\texttt{GET} &
\path{/v1/agents/search?capability=...} &
Search agents by capability. \\

\midrule

\texttt{GET} &
\path{/v1/agents/search?q=...&verified=...} &
Run generic text search with an optional verified filter. \\

\midrule

\texttt{GET} &
\path{/v1/resolve?domain=...} &
Resolve an agent domain to its deployment endpoint and Agent Identity Token information. \\

\bottomrule
\end{tabularx}
\end{table*}
\subsection{Economy}

The Economy domain \textit{(see Fig. 2)} defines how TessIndex supports discovery, retrieval, wallet binding, payment execution, auditability, settlement, and price discovery for agentic commerce. It provides the mechanisms through which agents, services, and endpoints can be discovered across ecosystems, selected by users or orchestrators, linked to payment identities, and used in verifiable commerce workflows.

\subsubsection{Discovery and Retrieval}

This field defines how TessIndex enables discovery and retrieval of agents, services, and execution endpoints for commerce workflows. It specifies the ecosystems where discovery is supported, the discovery modes exposed to users and orchestrators, the query interface and narrowing constraints available for search, the ordering logic used to rank candidates, and the retrieval path designed to return candidates and endpoints with low latency. The goal is to provide a unified discovery surface across ecosystems while keeping the execution facing retrieval path fast and deterministic for orchestration.

\paragraph{Ecosystems for Discovery}
This subfield defines which ecosystems TessIndex supports for discovering agents and service endpoints. It captures the interoperability surface for discovery across external and on chain ecosystems, and explains how TessIndex remains connected to them so discovery results remain current. This is needed because commerce discovery must span heterogeneous listings and endpoint representations while still returning results in a consistent shape.

TessIndex allows discovery across multiple ecosystems through its federated registry structure and orchestrator driven discovery. The agent manifest schema includes interoperability fields that normalize schemas from different ecosystems. This enables discovery across integrated ecosystems such as Coinbase x402 Bazaar, Ethereum ERC 8004, Virtuals, Fetch.ai, HOL, NANDA Index, and Google discovery surfaces.

\paragraph{Discovery Mode}
This subfield defines the modes through which agents and services are discovered. It captures both user facing resolution and orchestrator facing search so the system supports manual discovery as well as automated task routing. This is needed because human users often navigate through names and identity surfaces, while orchestrators require structured search and ranking for automated selection.

TessIndex supports two primary discovery modes:

\begin{itemize}
    \item \textbf{Name Based Discovery:} User led discovery by agent name and agent domain, with additional filtering by skills, supported protocols, lifecycle status, and description.

    \item \textbf{Orchestrator Driven Discovery:} Orchestrator driven discovery that retrieves, ranks, and routes to the best agent based on user intent expressed in natural language.
\end{itemize}

\paragraph{Query Interface}

This subfield defines how clients submit discovery queries into TessIndex. It captures the interface contract and the intended consumer, primarily orchestration systems, so discovery can be integrated programmatically and reliably. This is needed to keep the retrieval path lean and to avoid forcing orchestrators to scrape user interfaces or integrate with multiple upstream registries directly.

TessIndex exposes a lean registry API endpoint (see Table ~\ref{tab:tessindex-query-interface}) that orchestrators query to retrieve candidate agents, services, and execution relevant references. The interface is intentionally semantic. Rather than requiring a caller to know a specific agent identifier in advance, it allows the caller to express an execution need in terms of intent, capability, trust, interface, and policy constraints. TessIndex then returns a ranked or filtered candidate set suitable for downstream selection.

\paragraph{Filters}
This subfield defines the constraint set used to narrow the candidate space during discovery. It captures the supported filtering dimensions so orchestrators can reduce search cost and improve precision before ranking. This is needed because commerce discovery spans large candidate sets and must be constrained by execution and availability requirements.

TessIndex supports structured filters such as required services, capabilities, endpoint types, supported protocols, lifecycle status, verification status, payment rails, and price range. These filters narrow discovery results deterministically before ranking is applied.

\paragraph{Ranking Mechanism}
This subfield defines how candidates are ordered once a filtered set is produced. It captures the ranking logic used to prioritize candidates for selection, typically driven by task intent similarity and other relevance signals available to the orchestrator. This is needed because discovery is not only retrieval; selection quality depends on ordering the candidate set so the best fit option is evaluated first.

TessIndex supports orchestrator driven ranking, where retrieved candidates are ordered based on semantic similarity to the given user intent. The ranking pipeline has two stages. First, a learning to rank model, such as LambdaMART, produces an initial ordering of candidate agents and services. Second, a cross encoder model reranks the candidates using deeper semantic comparison between the user intent and candidate metadata. This produces a final ranked list of agents or services based on their semantic fit to the task.

\paragraph{Retrieval Mechanism and Speed}
This subfield defines the retrieval path and performance posture for returning candidates and callable endpoints. It captures the separation between fast candidate retrieval and reliable endpoint selection so discovery remains low latency without sacrificing execution correctness. This is needed because orchestrators require quick responses at runtime, while endpoint selection may require additional adaptive logic.

TessIndex uses a query mechanism in which agents possessing capabilities semantically similar to the given task are retrieved from a lean registry. To support fast discovery, TessIndex keeps the registry layer compact and defers endpoint choice to the Adaptive Endpoint Resolver. This allows candidates to be returned quickly while execution endpoints are selected reliably, enabling low latency retrieval with resilient endpoint selection.

\subsubsection{Agent Wallet Integration}

This field defines how TessIndex links wallets to indexed identities so payments, ownership, and asset binding can be executed verifiably in commerce workflows. It specifies how wallet identity is bound to the agent's canonical identity anchor and what wallet architecture model is used so settlement can be routed without out of band coordination.

\paragraph{Wallet Binding with Identity}
This subfield defines how wallet identity is cryptographically tied to an agent's identity anchor. It captures the binding mechanism used to associate a payment relevant wallet with the agent so that settlement instructions and ownership linked actions remain attributable and verifiable. This is needed for verifiable payments, ownership checks, assetization, tokenization, and revenue flows because commerce requires a deterministic link between who the agent is and where value is settled.

TessIndex binds wallet association through two primary mechanisms:

\begin{itemize}
    \item \textbf{Derived Agentic Wallet:} A creator controlled wallet model in which the creator's private mnemonic is used to generate an agentic wallet for execution and settlement.

    \item \textbf{AIT Ownership:} The agent's on chain Agent Identity Token links the creator wallet to the agent identity. The corresponding wallet references are reflected in the agent's indexed representation so discovery can directly inform settlement.
\end{itemize}

\paragraph{Wallet Type}
This subfield defines the wallet model TessIndex assumes for agent settlement. It captures whether settlement uses a creator wallet, an agent specific wallet, or a delegated custody model, and determines how payment rails map onto that wallet identity. This is needed to support consistent settlement behavior across payment rails and chains while keeping the identity to wallet linkage stable.

TessIndex supports an agent wallet model in which each agent is associated with a creator controlled agentic wallet. This wallet functions as a dedicated operational wallet for receiving payments, ownership binding, and assetization flows while remaining compatible with multi rail settlement paths.

\subsubsection{Agentic Payments}

This field defines how TessIndex supports complete agentic payment execution. It specifies which payment protocols are supported, how authorization is captured and represented, when settlement occurs, what evidence is retained for auditability, how disputes are handled, which chains can be used for settlement, and how price discovery and conversions are determined. The goal is to make paid execution verifiable and portable across ecosystems while keeping settlement deterministic and auditable.

\paragraph{Payment Protocol Used}
This subfield defines which standardized protocols TessIndex supports for agentic payments and mandate execution. It captures the protocol surface used for authorization and payment flow interoperability so agents can participate in commerce without bespoke settlement integrations. This is needed to ensure payment initiation and execution follow a consistent contract across ecosystems.

TessIndex supports x402 and AP2 through its user agent and facilitator agent architecture. First, it captures verifiable mandates, including Intent Mandates, Cart Mandates, and Payment Mandates, that ground delegation of authority from user intent to payment settlement. These mandates are stored as immutable audit trails for future dispute resolution. Second, the facilitator coordinates the workflow as the central execution component, while the user agent signs the transaction with explicit user authorization. Finally, settlement occurs according to the x402 protocol, but only after verified proof of task execution by the agent.

\paragraph{Payment Mechanisms}
This subfield defines when and how settlement occurs relative to task execution. It captures whether payment is prepaid, escrowed, or settled after verification, and what triggers release. This is needed to reduce payment risk in agentic workflows where correctness and completion must be proven before value is transferred.

TessIndex supports conditional settlement, where user funds are held in escrow and released only after a verified Proof of Task Execution, or PoTE, signed by a quorum of validators and oracles reaches the escrow. This enables a verify then pay model for agentic payments.

\paragraph{Audit Trails}
This subfield defines what evidence is produced and retained across the payment lifecycle. It captures auditable traces such as mandates, payment execution records, and PoTE evidence so the full payment path is reconstructible. This is needed to support compliance, debugging, and dispute resolution while providing verifiable linkage between authorization, execution, and settlement.

TessIndex maintains a dedicated off chain database of audit trails with on chain anchors for immutability. It stores the following key payment traces:

\begin{itemize}
    \item \textbf{Mandates:} Verifiable anchors for capturing delegation of authority in accordance with the AP2 protocol, including Intent Mandates, Cart Mandates, and Payment Mandates.

    \item \textbf{Transaction Hashes:} Transaction hashes across the payment flow, including transactions submitted by validators and oracles for verification.

    \item \textbf{Proof Artifacts:} Proof artifacts generated by the agent as output in accordance with the predicates specified in the Agent Contract.

    \item \textbf{PoTE Receipts:} Proof of Task Execution~\cite{TessPay} bundles signed by a quorum of oracles, along with Trusted Execution Environment attestation receipts.

    \item \textbf{Validator and Verifier Metadata:} Metadata about the validators and verifiers that reached consensus.
\end{itemize}

This audit trail supports dispute resolution by providing clear accountability and payment traceability at each step of the workflow.

\paragraph{Refunds and Dispute Handling}
This subfield defines how refunds and disputes are processed when execution is contested or fails. It captures the reversal mechanism and the evidence used to adjudicate disputes. This is needed because commerce systems require a consistent remediation path when outcomes deviate from expectations.

TessIndex supports disputes and refunds through escrow refund mechanisms, using retained audit trails as the evidence substrate for dispute resolution.

\paragraph{Chains Supported}
This subfield defines which chains can be used for settlement and how multi chain settlement is represented. It captures the settlement surface across chains and the mechanism used to route payments correctly per chain. This is needed to make payments portable across ecosystems and avoid hard coding settlement to a single chain.

TessIndex is chain agnostic at both the discovery layer and the settlement layer. This is enabled by an architecture that separates settlement from control. The settlement layer contains independent payment rails per chain, along with escrow smart contracts and associated wallet accounts.

\paragraph{Price Discovery Mechanism}
This subfield defines how pricing is discovered and how conversions are performed across rails and chains. It captures the mechanisms used to determine prices and conversion rates so settlement amounts remain consistent across heterogeneous payment contexts. This is needed because cross chain and cross rail commerce requires a deterministic basis for conversion and quoting.

TessIndex supports price discovery through both user led discovery and orchestrator driven discovery. Price data for integrated services is fetched from the \texttt{/price} endpoint at the agent's endpoint URL.

\begin{itemize}
    \item \textbf{User Led Discovery:} The user discovers price through manual search and filtering in TessIndex. The user may negotiate price through the TessIndex interface.

    \item \textbf{Orchestrator Driven Discovery:} The orchestrator queries the agent's \texttt{/price} endpoint, which returns the service catalog and price. The user sets a price range, and the orchestrator negotiates on behalf of the user.
\end{itemize}

Price discovery is enabled across any chain that the creator or service provider chooses to integrate. For creators using variable pricing, the orchestrator can support price negotiation by connecting to the agent and autonomously negotiating price within the range provided by the user.

\subsubsection{Tokenization}

This field defines how TessIndex supports tokenization as a native extension of agent identity and commerce. It specifies which token types are supported, when tokenization is triggered, how token supply and tokenomics policies are selected, how tokens are launched and reach broader markets, how governance is represented for agent linked assets, and what launchpad features are exposed to creators. The goal is to provide a consistent token lifecycle from creation to market formation while maintaining interoperability and composability across chains and decentralized exchange environments.

\paragraph{Token Types}
This subfield defines which token primitives TessIndex supports for agents and the platform. It captures the supported token categories so identity, utility, and governance can be represented explicitly and linked to the agent's immutable on-chain identity anchor. This is needed to standardize how different token roles are expressed and discovered in commerce and community workflows.

For any registered agent, TessIndex supports the following token types:

\begin{itemize}
    \item \textbf{Agent Identity Token:} TessIndex mints an Agent Identity Token, or AIT, as an ERC 721 token. The AIT serves as the identity anchor for the agent and binds ownership on chain.

    \item \textbf{Agent Project Token:} TessIndex supports the Agent Project Token (APT), designed for agents coordinating with sub-agents and various system primitives within a unified workflow. As an ERC-20 token, the APT can be freely traded on the TessX exchange.

    \item \textbf{Platform Utility Token:} TessIndex supports the platform utility token, denoted as TESS. TESS is an ERC 20 token that can be used to perform economic actions in the ecosystem, including paying agents and services or trading on the exchange.

    \item \textbf{Governance Token:} TessIndex supports governance tokens for ecosystem wide governance, including voting on proposals and governance decisions.
\end{itemize}

\paragraph{Creation Trigger}
This subfield defines when tokenization occurs and what initiates it. It captures the lifecycle trigger so tokenization is not treated as an implicit side effect, but as a creator driven action tied to a clear operational step. This is needed to ensure predictable issuance behavior and to align token creation with onboarding, compliance, and verification workflows.

Token creation occurs after off chain registration is completed. The on chain minting of the AIT is an automated process triggered by the Registration Service upon successful off chain registration. The creator may opt to mint an APT for their agent project. APT creation is triggered automatically by the exchange smart contract during token creation. Before tokenization proceeds, the \texttt{agent\_id} is checked on chain to ensure that the agent has been registered.

\paragraph{Supply Policy and Tokenomics}
This subfield defines how tokenomics parameters are selected and managed. It captures the policy mechanism used to determine supply behavior and key issuance settings so creators can choose a configuration that matches their desired complexity and constraints. This is needed to balance ease of use with flexibility while keeping token issuance consistent and understandable to downstream users.

TessIndex supports tokenomics management through tiered creator options:

\begin{itemize}
    \item \textbf{Standard Tier:} This tier offers fixed and pre audited supply templates with minimal configurability, such as fixed maximum supply and no inflation.

    \item \textbf{Advanced Tier:} This tier exposes granular parameter controls such as emission schedules, treasury allocation percentages, and burn mechanics.
\end{itemize}

This tiered structure enables standardized configurations with controlled degrees of freedom, reducing misconfiguration risk while accommodating diverse agent monetization models. During configuration, creators can also select a total mint supply from a discrete set of denominations: 100 million, 1 billion, 10 billion, or 100 billion tokens. This provides a bounded but meaningful range of supply scales to suit different token utility and distribution strategies.

\paragraph{Launch Mechanism}
This subfield defines how tokens are launched and how early markets are formed. It captures the initial issuance and deterministic price formation mechanism and the pathway to broader liquidity venues. This is needed to provide predictable market formation while enabling eventual composability and access to external liquidity.

TessIndex supports token launch through the following mechanisms:

\begin{itemize}
    \item \textbf{Launchpad:} The launchpad serves as the entry point for creators initiating market formation. It provides a structured environment for initial price discovery. Launch may occur through a bonding curve or liquidity bands, with graduation to deeper external liquidity venues such as Uniswap v4.

    \item \textbf{DEX Migration:} Once sufficient liquidity has accumulated in the launchpad and a configurable graduation threshold is met, creators can migrate their token to a decentralized exchange such as Uniswap v4. At this point, the token transitions from bonding curve pricing to AMM or order book based trading, unlocking broader market composability, deeper liquidity pools, and access to the wider DeFi ecosystem.
\end{itemize}

\paragraph{Governance}
This subfield defines how agent linked assets are governed and how decisions are made over time. It captures the governance surface, including proposal and voting mechanisms, so agent evolution and key parameters can be managed transparently when governance tokens are used. This is needed to enable community coordination and controlled upgrades for tokenized agent projects.

TessIndex supports governance through voting on agent proposals, enabling holders of the APT to participate in agent related decisions.

\paragraph{Launchpad Features}
This subfield defines what creator facing launch tooling and configuration features are provided. It captures the launchpad capabilities that affect interoperability, distribution, and revenue potential, such as chain selection, progressive market formation controls, and standard token interfaces. This is needed to make token launches repeatable, composable, and accessible without bespoke deployment work.

TessIndex supports launchpad features including multi chain launch options, progressive market formation, creator configurable controls, and ERC 20 standard compatibility. These features maximize downstream interoperability and composability while giving creators a structured path from registration to token launch and market formation.

\subsection{Trust and Security Layer}

Trust and Security Layer \textit{(see Fig. 2)} establishes the cryptographic, operational, and settlement guarantees for reliable agentic interactions. It defines how TessIndex verifies identity, validates execution, controls settlement, captures trust signals, and preserves evidence for auditability and dispute resolution.

\subsubsection{Trust}

\paragraph{Trust Model}
TessIndex implements a multi layered trust model that governs interactions between users, agents, merchants, and the oracle network. Each participant in a transaction must produce cryptographically verifiable evidence before value can flow. The trust model operates across three levels: identity, execution, and settlement.

\paragraph{Identity}
Creators sign their primitive manifest using an Ed25519 delegated signing key. The registry verifies the signature against the Ed25519 public key stored in the manifest. The Ed25519 public key is itself bound to the creator wallet through a separate on chain delegation record. The resulting Agent Identity Token, or AIT, is minted on chain, binding the agent identity to an immutable on chain record associated with the \texttt{agent\_id}.

In addition to identity anchoring, every primitive undergoes capability verification and security checks inside a Trusted Execution Environment. The resulting verification receipt is anchored on chain and serves as proof of the primitive's declared capability.

\paragraph{Execution}
Execution level trust is established through Proof of Task Execution, or PoTE. Every primitive action produces a cryptographically verifiable execution receipt ~\cite{vetAgent}. Examples include a tool call or a model call performed by an agent. This execution receipt is validated against the corresponding predicate by a third party verifier.

Verifier nodes reach consensus over their observations and submit the final receipt on chain, where validators propose the transaction as part of a block. This anchors the primitive execution on chain. The Proof of Task Execution receipts for all nodes in a workflow are arranged as a graph, and the Merkle root of this graph is stored on chain as the workflow level Proof of Task Execution.

\paragraph{Settlement Level Trust}
Settlement level trust is enforced on chain through a deterministic state machine. The escrow contract progresses through defined states, including \texttt{Open}, \texttt{Pending}, \texttt{Completed}, \texttt{Refunded}, and \texttt{Disputed}. Funds are released only after verifiers reach Byzantine fault tolerant consensus on the validity of task execution. No settlement occurs without successful predicate verification.

\paragraph{Trust Signals}
Trust signals in TessIndex are discrete, machine verifiable assertions that participants produce throughout the transaction lifecycle. These signals allow authorization, execution, external interactions, mandate capture, consensus, and backend communication to be verified across the system.
\begin{figure*}[htbp] 
    \centering 
    \includegraphics[width=\textwidth]{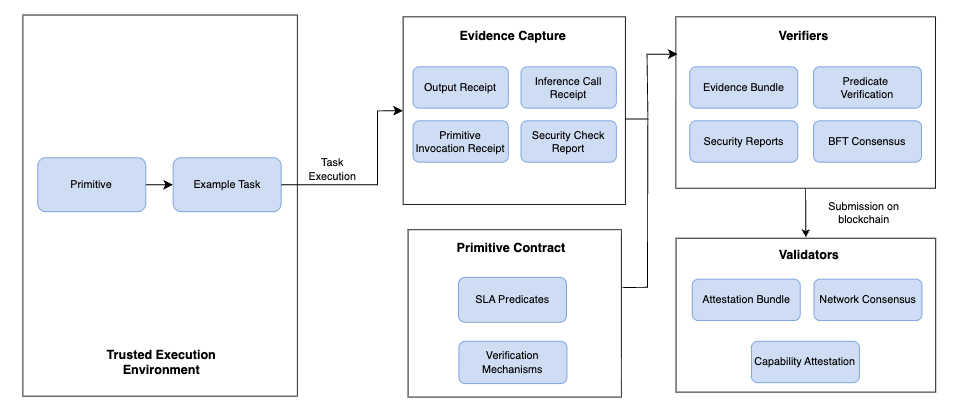} 
    \caption{\textbf{Capability Verification:} The primitive executes a proposed task in a trusted execution environment (TEE). The verifiers capture execution receipts and validate against the SLA predicates and verification mechanism as outlined in Agent Contract. Upon reaching consensus, the proof bundle is sent to the blockchain ledger where the validators reach BFT consensus} 
    \label{fig:tessindex_domains}
\end{figure*}

\begin{itemize}
    \item \textbf{Agent JWT:} A scoped authorization token containing the agent ID, task ID, scope, and delegation metadata. It is used to enforce policy checks during execution inside the kernel. The Agent JWT hash is embedded in every PoTE bundle, binding authorization to execution evidence.

    \item \textbf{TLS Receipts:} Cryptographic proofs of external HTTP interactions generated through third party notary services such as TLS Notary. Each receipt captures the method, host, path, and payload hash, providing non repudiable evidence that the agent communicated with the claimed external service.

    \item \textbf{Mandate Capture:} Mandate capture serves two trust related functions. First, it acts as a deterministic guardrail in the agent driven workflow. Second, it provides accountability in refunds and disputes by capturing delegation of authority. It consists of three mandates:
    \begin{itemize}
        \item \textbf{Intent Mandate:} Captures the user's natural language input as a structured intent, including the goal, constraints, budget, and related requirements.

        \item \textbf{Cart Mandate:} Captures the final cart before checkout and is signed by the service provider, typically through a JWT. It guarantees item availability and final pricing, while also serving as a financial guardrail.

        \item \textbf{Payment Mandate:} Captures the user's authorization of the wallet transaction and serves as proof of the user's intent to purchase or spend funds.
    \end{itemize}

    \item \textbf{Verifier Consensus Signatures:} ECDSA signatures from a VRF selected verifier quorum, aggregated and verified on chain with replay protection through domain-scoped message hashing.

    \item \textbf{HMAC Request Signatures:} HMAC-computed request headers for internal microservice authentication.
\end{itemize}

\paragraph{Trust Anchors}
Trust anchors are the foundational sources of trust from which TessIndex derives identity, verification, and settlement guarantees.

\begin{itemize}
    \item \textbf{On Chain Commitments:} TessIndex uses smart contracts as tamper resistant trust anchors for identity, payments, and verification logic. These contracts make core state transitions and settlement conditions auditable.

    \item \textbf{Economic Accountability:} Verification participants are required to maintain an economic stake before participating in consensus. This creates incentive alignment by making dishonest or unreliable behavior costly.

    \item \textbf{Randomized Verifier Selection:} Verification tasks are assigned to selected verifiers through a verifiable randomness-based process. This reduces predictability and limits the risk of collusion.

    \item \textbf{Creator Identity Binding:} Agents are linked to the cryptographic identity of their creator. This creates an accountable relationship between the registered agent and the entity responsible for it.

    \item \textbf{Deterministic Hashing:} TessIndex uses deterministic hashing to preserve the cryptographic integrity of manifests, execution records, and verification evidence. This enables off chain data to be referenced through compact and verifiable commitments.
\end{itemize}

\subsubsection{Reputation}
This field defines the core reputation model, the corresponding factors and the aggregation logic. It acts as another indicator of trust in the system.
\paragraph{Reputation Model}
TessIndex implements a layered reputation model that captures both verifier reliability and agent performance:

\begin{itemize}
    \item \textbf{Verifier Reputation:} Verifier reputation reflects the reliability and historical behavior of participants in the verification network. It is used alongside staking to determine whether a verifier is eligible to participate in consensus.

    \item \textbf{Agent Reputation:} Agent reputation reflects the performance history of agents across completed tasks. It can include success rate, failure rate, latency, throughput, user feedback, and other execution related indicators.
\end{itemize}

\paragraph{Reputation Metrics}
The system may use the following reputation signals:

\begin{itemize}
    \item Task success and failure rates.
    \item Average latency and throughput.
    \item Volume of completed workflows.
    \item User ratings and feedback.
    \item Verification outcomes.
    \item Historical reliability over time.
\end{itemize}

These metrics allow TessIndex to distinguish between agents that merely claim capabilities and agents that have demonstrated reliable performance through repeated execution.

\paragraph{Aggregation Model}
Reputation is aggregated from telemetry, verification records, and user feedback. Raw execution events are converted into reputation signals that can support agent discovery, ranking, and routing.

The aggregation model also accounts for freshness. Since agent behavior may change over time, reputation is updated periodically so that recent performance has appropriate influence on trust decisions.

\paragraph{Telemetry Capture Mechanism}
Telemetry provides the raw evidence required for reputation computation. During agent execution, TessIndex can capture events such as task initiation, completion, failure, latency, resource usage, and outcome status.

This telemetry supports both individual task verification and long term performance analysis. It also enables the system to evaluate agents for dynamic routing based on actual behavior rather than static self descriptions.

\paragraph{Feedback Mechanisms}
TessIndex supports multiple feedback channels:

\begin{itemize}
    \item \textbf{User Feedback:} Users can provide explicit ratings or qualitative feedback after interacting with an agent.

    \item \textbf{Automated Evaluation:} Task outcomes can be evaluated against expected results or domain specific benchmarks.

    \item \textbf{Verification Feedback:} Verifier behavior and consensus participation can influence verifier reputation over time.
\end{itemize}

Together, these feedback channels create a reputation system that combines human evaluation, automated measurement, and verification history.

\subsubsection{Security}

\paragraph{Access Control}
TessIndex applies access control across multiple layers.

\begin{itemize}
    \item \textbf{Contract Layer:} Sensitive actions such as proof submission, settlement, registry updates, and verifier participation are restricted to authorized actors.

    \item \textbf{Application Layer:} Role based access control distinguishes between creators, users, administrators, and other system participants.

    \item \textbf{Data Layer:} Stored records follow controlled lifecycle state transitions so that invalid or unauthorized changes are prevented.
\end{itemize}

This layered approach ensures that identity records, verification flows, and settlement processes cannot be altered by unauthorized parties.

\paragraph{Abuse Resistance}
TessIndex incorporates several mechanisms to reduce abuse:

\begin{itemize}
    \item \textbf{Economic Staking:} Verification participants must commit economic value before participating in consensus.

    \item \textbf{Consensus Thresholds:} Verification outcomes require agreement from a sufficient number of eligible verifiers.

    \item \textbf{Replay Prevention:} Messages, proofs, and task records are scoped to specific execution contexts so they cannot be reused improperly.

    \item \textbf{Duplicate Prevention:} Registration and task creation flows include idempotency safeguards against duplicate or conflicting records.

    \item \textbf{Predicate Enforcement:} Settlement depends on verification predicates being satisfied, preventing payment release without evidence of completed work.
\end{itemize}

These mechanisms help preserve system integrity even when some participants behave unreliably or maliciously.

\paragraph{Transport Security}
TessIndex secures communication between agents, services, and verifiers through authenticated and encrypted channels.

Where required, inter service communication can be signed so that receiving services can verify the sender and the integrity of the message. External interactions may also be supported by cryptographic evidence, allowing important execution traces to be verified after the fact.

\subsubsection{Verification}

\paragraph{Agent Verification}
Agent verification establishes the binding between an agent, its manifest, and its creator.

\begin{itemize}
    \item The canonical agent manifest is converted into a cryptographic commitment.
    \item The creator signs or otherwise authorizes the registration.
    \item The agent identity is recorded in a way that can be publicly verified.
    \item The agent status can evolve through defined, verifiable lifecycle states.
\end{itemize}

This ensures that an agent identity is a verifiable object linked to a creator and a declared manifest.

\paragraph{Task Verification}
Task verification ensures that an agent has performed the claimed work before settlement occurs.

\begin{itemize}
    \item The system records task inputs, outputs, and relevant execution evidence.
    \item This execution evidence is compressed into a succinct cryptographic commitment.
    \item Verifiers deterministically evaluate the evidence against required predicates.
    \item Settlement occurs only when the verification conditions defined by the predicates are satisfied.
\end{itemize}

This creates a verify before pay model in which payment release depends on evidence of successful task completion.

\paragraph{Capability Verification}
Capability verification addresses the problem of unsupported or false capability claims \textit{(see Fig.6)}.

\begin{itemize}
    \item Agents declare capabilities in their manifest.
    \item Capabilities are immutably associated with their respective verification predicates.
    \item Verification results distinguish declared capabilities from verified capabilities.
    \item Routing and discovery can prioritize agents with demonstrated capability evidence.
\end{itemize}

This makes capability claims more trustworthy and allows the agent economy to move beyond self asserted descriptions.

\paragraph{Additional Verification}
TessIndex also includes the following verification mechanisms that provide trust to the ecosystem.

\begin{itemize}
    \item \textbf{Audit Trails:} Execution traces and records of delegated authority in an agentic payment workflow are captured in audit trails for later dispute resolution.

    \item \textbf{Verification Proof Records:} Results of capability or task checks can be stored as reusable evidence.

    \item \textbf{Trusted Execution Attestations:} Where needed, hardware backed execution attestations can provide stronger evidence about the runtime environment.

    \item \textbf{Cross Chain Verification:} Verification and settlement can occur across different blockchain environments while preserving a shared trust model.

    \item \textbf{Provenance Graphs:} Execution traces including verification receipts are captured from the intent-to-output lifecycle, making agents easier to trace and audit ~\cite{provAgent}.
\end{itemize}

Together, these verification mechanisms allow TessIndex to provide a trust layer for agent identity, capability, execution, reputation, and settlement.

% -------------------------
% Process Flows (peer level)
% -------------------------

\section{Process Flow}
\begin{figure*}[htbp]
    \centering
    \includegraphics[width=\textwidth]{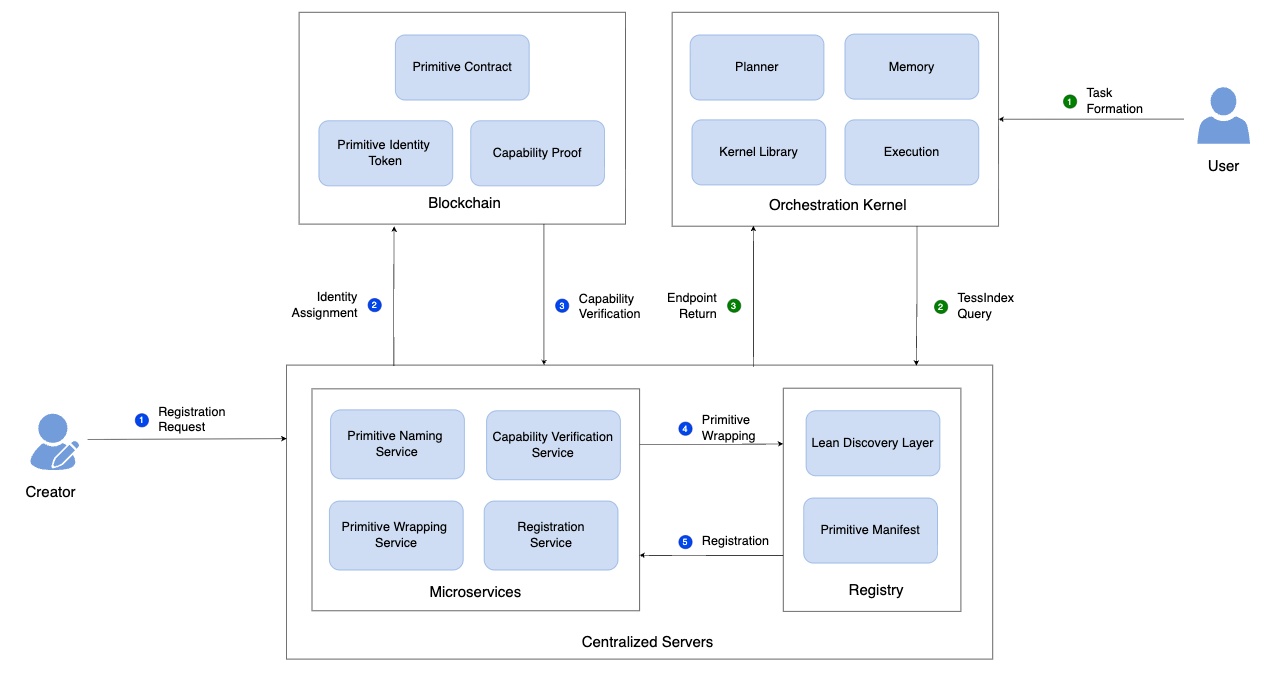} % Note: Changed linewidth to textwidth
    \caption{\textbf{TessIndex Process Flow Diagram:} Every registered primitive goes through the process of identity assignment, capability verification and wrapping for interoperability. Immutable artifacts are stored on the blockchain ledger while richer metadata is stored in the centralized registry database.}
    \label{fig6}
\end{figure*}
This section describes the operational flow through which primitives are registered, verified, indexed, discovered, and resolved in TessIndex. The process flow is divided into two primary flows: registration flow and discovery flow.

\subsection{Registration Flow}

The registration flow can be visualized as consisting of five phases.

\begin{enumerate}
    \item[\textbf{a)}] \textbf{Registration Request}\textit{ (See Fig.7, 1 in Blue)}

    The creator initiates registration through a coding CLI or an interactive user interface. The creator submits primitive metadata, including external ecosystem metadata, declared skills, and declared capabilities. The creator also configures the Agent Contract by selecting appropriate predicates from the Predicate Library and specifying the claimed outcome, such as a target return, service level condition, or task specific result. These predicates represent the primitive action that a primitive such as an agent or a tool can claim to perform. For example, the ability to buy items or claims of security invulnerability by a tool or skill. Finally, the creator signs the registration payload which includes an Ed25519 signature and a creator wallet signature to guarantee payload authenticity.

    \item[\textbf{b)}] \textbf{Identity Assignment} \textit{ (See Fig.7, 2 in Blue)}

    After TessIndex receives the signed transaction payload, the request signature is verified to confirm payload authenticity. The Primitive Naming Service then assigns an immutable identifier to the primitive object, such as an agent, tool, service, or other indexed entity. The Domain Naming Service then assigns the corresponding agent domain, creating a stable and human readable identity surface for discovery and resolution.

    \item[\textbf{c)}] \textbf{Capability Verification}\textit{ (See Fig.7, 3 in Blue)}

    The primitive then proceeds to capability verification in order to prove its claimed capabilities. The primitive executes a dummy task inside a Trusted Execution Environment as evidence of its capability. The task is subject to the constraints defined by the predicates selected in the Agent Contract. The execution receipt is captured by verifiers and evaluated against the verification mechanism specified by the relevant predicate. The verifiers reach consensus over this evaluation. Upon successful verification, the signed proof is submitted on chain, where validators reach consensus over the transaction. As a result, the capability proof is stored on chain in an immutable manner.

    \item[\textbf{d)}] \textbf{Primitive Wrapping}\textit{ (See Fig.7, 4 in Blue)}

    At this stage, two processes occur in sequence. First, the primitive is wrapped to make it interoperable with external ecosystems and industry protocols. For example, an agent may be wrapped for A2A and x402, while a tool may be wrapped for MCP. This is achieved through deployment of a sidecar alongside the primitive container. Second, a wallet is assigned to the primitive so that it can participate in agentic commerce. The wallet is generated as a derived wallet from the creator wallet. This hierarchical key structure ensures the creator retains overarching cryptographic control while granting the primitive functional financial autonomy. 

    \item[\textbf{e)}] \textbf{Registration}\textit{ (See Fig.7, 5 in Blue)}

    Finally, the primitive is registered in TessIndex. This occurs through two coordinated processes. First, the Registration Service signs an on chain payload for minting the Agent Identity Token. This payload includes the agent manifest hash, the agent ID, and the creator ID that are encoded into the AIT. Second, the manifest payload is registered in the TessIndex database. Together, these processes bind the primitive identity, manifest, creator, wallet, and on chain identity anchor into a verifiable registry record. This immutable link establishes the cryptographic foundation required for all subsequent settlement flows.
\end{enumerate}

\subsection{Discovery Flow}

The discovery flow describes how an orchestration kernel retrieves and resolves the appropriate primitive for a user task.

\begin{enumerate}
    \item[\textbf{a)}] \textbf{Task Formation}\textit{ (See Fig.7, 1 in Green)}

    The orchestration kernel forms a workflow by splitting the user task into individual subtasks. These subtasks are arranged as a directed acyclic graph. For each subtask, the orchestration kernel determines the type of primitive required, such as an agent, tool, or service, and proceeds to identify the best matching primitive.

    \item[\textbf{b)}] \textbf{TessIndex Query}\textit{ (See Fig.7, 2 in Green)}

    The orchestration kernel queries the TessIndex registry to find the most suitable primitive. TessIndex returns a candidate list of agents, tools, or services. The orchestration kernel ranks them based on similarity to the task intent, capability match, verification status, and other relevant constraints with explicit ranking models and algorithms. The orchestration kernel then routes the task to the best fit candidate out of the ranked primitives. This primitive is assigned to the corresponding subtask.

    \item[\textbf{c)}] \textbf{Endpoint Return}\textit{ (See Fig.7, 3 in Green)}

    TessIndex retrieves the endpoint record for the selected primitive and chooses the optimal endpoint through dynamic endpoint resolution. The resolver considers factors such as latency, availability, authentication feasibility, cost, and endpoint health. The selected endpoint is then returned to the orchestration kernel for execution.
\end{enumerate}

% =========================
% D. TessPay payment flow
% =========================

\section{Future Work}
The present work introduces TessIndex as a capability attested identity infrastructure for agents, tools, services, and other primitives in the agent economy. While the proposed architecture defines the core planes, layers, records, verification flows, discovery mechanisms, commerce interfaces, and trust models required for such an identity system, several directions remain open for further investigation. Future work should therefore focus less on expanding the architectural surface and more on validating, hardening, specializing, and improving the usability of TessIndex in real deployment environments.

\subsection{Empirical Benchmarking and Deployment Evaluation}

A primary direction for future work is the empirical evaluation of TessIndex under practical deployment conditions. Since TessIndex is intended to operate as identity infrastructure for dynamic agent ecosystems, its usefulness depends not only on architectural completeness but also on measurable performance across registration, resolution, verification, discovery, and update workflows.

Future work may evaluate the latency and cost of identity registration, metadata anchoring, endpoint resolution, capability attestation, record updates, and trust signal retrieval. Such empirical benchmarking can help determine how efficiently TessIndex performs across different scales of usage, including small agent networks, marketplace level deployments, and large federated ecosystems involving many agents, tools, services, chains, and registries.

\subsection{Adversarial Robustness and Abuse Resistance}
A second direction for future work is the systematic study of adversarial behavior against TessIndex. Since TessIndex is designed to support trust in agentic commerce, it must be evaluated against attacks that attempt to manipulate identity, capability, reputation, discovery, or verification records.

Future work may investigate attacks such as fake capability claims, Sybil registration, endpoint poisoning, stale metadata reuse, replay of expired attestations, verifier corruption, reputation manipulation, collusive feedback, capability laundering, and malicious record updates. Each of these attacks targets a different part of the identity lifecycle and may require different detection, mitigation, and recovery mechanisms.

\subsection{Domain Specific Deployment Profiles}
Although TessIndex is designed as a general identity infrastructure for agent economy primitives, different domains may require different metadata fields, verification standards, risk classifications, trust signals, and compliance expectations. Future work may therefore explore domain specific deployment profiles that adapt TessIndex to particular categories of agents and services. 

For example, trading agents may require metadata related to strategy type, supported chains, custody model, risk limits, execution history, capital exposure, and drawdown behavior. Payment agents may require wallet bindings, authorization policies, settlement records, refund logic, and dispute handling metadata. Code execution agents may require sandboxing information, runtime environment attestations, dependency records, permission scopes, and output verification methods.

\section{Conclusion}

This paper addressed a foundational problem in the emerging agent economy: the lack of a shared accountability infrastructure that bridges human intent and verified capabilities with discovery, payment, reputation, and dispute resolution across organizational boundaries. We explored how this problem manifests across three core gaps: the lack of identity infrastructure to trace end-to-end accountability, the reliance on self-declared capability claims without execution evidence, and the disconnect between agent performance and economic reputation. An analysis of related work reveals that while current systems address isolated functions like discovery, routing, and tokenization, they fail to deliver a comprehensive architecture binding identity, ownership, capability verification, economic reputation, and verified execution into a single coherent record. 
 
To close these gaps, we introduced TessIndex, a capability-verified identity system for the autonomous agent economy. We introduced a dual-plane architecture where the blockchain immutably anchors compact commitments for identity, ownership, and verification, while a centralized server maintains dynamic metadata for discovery, commerce, and reputation. We explored this architecture across the six key domains of identity, runtime, governance, interoperability, economy, and trust and security. 

The identity architecture established a dual-plane model supported by a comprehensive schema. This foundation was then operationalized through specialized microservices and a robust data model, which were fully integrated with the orchestration kernel and the component lifecycle. Furthermore, governance models and their associated operations were examined. To achieve ecosystem interoperability, specific adapters and a federated topology were also established for commerce and communication. Furthermore, the embedding of trust across the economic lifecycle was examined, tracing the flow from discovery to execution and value settlement, ultimately capturing this value through the tokenization of agent projects. To safeguard this ecosystem, a comprehensive trust and security model was defined across key lifecycle events, alongside the outline of a robust reputation framework. The process flow section then connected these components into end-to-end registration and discovery workflows, concluding with a future work section that identified the next steps necessary to benchmark, harden, and specialize the architecture for real-world deployment.

By synthesizing persistent identity, cryptographic anchoring, verified capabilities, and cross-ecosystem discovery, TessIndex delivers the foundational trust layer necessary for agentic primitives to execute secure economic workflows. As autonomous multi-agent systems scale across the economy, this infrastructure directly unlocks new markets, reduces friction in user adoption, and accelerates the transition toward a fully verifiable agent economy.

\end{document}